\documentclass[journal]{IEEEtran}
\usepackage{amsmath,amsfonts}
\usepackage{array}
\usepackage[caption=false,font=footnotesize,labelfont=rm,textfont=rm]{subfig}
\usepackage{textcomp}
\usepackage{stfloats}
\usepackage{url}
\usepackage{verbatim}
\usepackage{graphicx}
\usepackage{cite}
\usepackage[linesnumbered,ruled,vlined]{algorithm2e}
\usepackage{times}
\usepackage{epsfig}
\usepackage{}
\usepackage{amssymb}
\usepackage{hhline}
\usepackage{bm}
\usepackage{multirow}
\usepackage{comment}
\usepackage{indentfirst}
\usepackage{subfig} 
\usepackage{float}
\usepackage[pagebackref=true,breaklinks=true,colorlinks,bookmarks=false]{hyperref}

\begin{document}

\title{Non-exemplar Class-incremental Learning by Random Auxiliary Classes Augmentation and Mixed Features}
\author{Ke Song, Guoqiang Liang, Zhaojie Chen, Yanning Zhang,~\IEEEmembership{Senior Member, IEEE}
}
 
\maketitle
 
\begin{abstract}
Non-exemplar class-incremental learning refers to continual classifying of new and old classes without storing samples of old classes. Since only new class samples are available, catastrophic forgetting of old knowledge often occurs. In this paper, we propose an effective non-exemplar method called RAMF consisting of \textbf{R}andom \textbf{A}uxiliary classes augmentation and \textbf{M}ixed \textbf{F}eatures. On the one hand, we design a novel random auxiliary classes augmentation method, where one augmentation is randomly selected from three augmentations and applied to inputs to generate augmented samples and extra class labels. By extending the data and label space, the model can learn more diverse and transferable representations, which can prevent the model from being biased towards learning task-specific features and facilitate the transfer among different tasks. In a word, when learning new tasks, the random auxiliary class augmentation will reduce the change of feature space and improve model generalization. On the other hand, we propose to replace the new features with mixed features for model optimization since only using new features will largely affect the previous representation embedded in the old feature space. Instead, by mixing new and old features, the cosine similarity is improved by reducing the angle between the current and old features, which allows for better stability over long-term incremental learning without increasing the computational complexity. We have conducted extensive experiments on three benchmarks CIFAR-100, TinyImageNet and ImageNet-Subset, where our method outperforms the state-of-the-art non-exemplar methods and is comparable to high-performance replay-based methods.
\end{abstract}


\section{Introduction}
Current deep neural networks have achieved great success in various fields \cite{objdectionTCSVT, FSLTCSVT}, even surpassing human capabilities, but this success relies on large amounts of static training data. However, in real world, data usually arrives in a dynamic stream. It will consume expensive computational expenditure to iteratively retrain models to integrate new knowledge. Thus the model should be updated over time \cite{streams2023}. Therefore, incremental learning, also known as continual learning or lifelong learning, has attracted extensive interest recently. It refers to a model which can keep learning new tasks as the distribution of training data changes. Its main challenge is how to learn new tasks while maintaining performance on previously learned tasks \cite{2022NIPSstudy,cf1}.

As stated in previous works \cite{cl2, 2022PAMIsurvey}, class-incremental learning is the most challenging scenario, where the model needs to learn new classes incrementally. Most current class-incremental learning methods with high performance are based on sample replay \cite{AAERTCSVT, AdaptiveTCSVT}, which needs memory to store a few samples from old classes when new tasks arrive. Although storing samples can alleviate catastrophic forgetting effectively, it is not practical in some scenarios where privacy is a concern. Besides, the requirement for ever-increasing storage limits their application in long-term incremental learning \cite{TPAMISurvey}.

Due to these limitations, the influence of non-exemplar class-incremental learning (NECIL) is on the rise. Early NECIL works \cite{lwf,lwm,ewc} adopt various forms of distillation to preserve old knowledge. However, since a large number of samples for new classes and very few prototypes for old classes are simultaneously fed to the model, the model will occur a severe bias towards new classes \cite{WA, Bic}. Thus, these approaches still exist catastrophic forgetting of old knowledge. To mitigate the model bias, class augmentation \cite{il2a, pass} has recently been introduced to NECIL. Through spectral decomposition analysis \cite{il2a}, they discovered learning from a \textit{wider variety of classes} can enable the model to acquire \textit{transferable and diverse features}. These features can facilitate learning across different tasks and prevent the feature space from becoming cluttered with task-specific features, which causes bad model initialization and affects the learning of subsequent tasks. Despite the promising performance, these class augmentation methods for NECIL exist some drawbacks. For instance, the extra classes produced in \cite{il2a} contain too many mixed features, which makes the model hard to learn features of original samples. \cite{pass} generated a few auxiliary classes containing a relatively limited quantity of transferable features. On the other hand, to keep the representations of the old class stable in feature space, current approaches usually employ knowledge distillation to reduce feature variation between the current and previous feature extractor \cite{SemanticTCSVT}. However, since the models are continually updated by new features that drifted from old ones, the errors will accumulate, ultimately leading to a large change of feature space \cite{WA}.

To solve the above problems, we propose a new NECIL method RAMF consisting of \textbf{R}andom \textbf{A}uxiliary class augmentation and \textbf{M}ixed \textbf{F}eatures. In the former, three augmentation methods are randomly used in the initial stage to generate augmented samples and extra class labels. The many augmented classes can enable the model to learn sufficient transferable features to facilitate learning across different tasks. Meanwhile, since the auxiliary classes contain all original features, they can avoid feature space deviation from the original features. Moreover, the \textit{randomness} introduced by the auxiliary class mimics the incremental learning paradigm, which allows the model to learn `future' tasks at the initial stage. This can mitigate the model forgetting when encountering actual unseen classes in incremental stages. Second, we propose mixed features to replace new features to prevent the model from being over-optimized by too many new features. Compared with new features, the mixed feature has a larger similarity with the old feature. In other words, during the model update, the feature drift can be reduced, so the model can still extract discriminative features for old classes. Besides, by assigning different weights to old features and new features, the mixed features are better adapted to continual learning scenarios under different data augmentation than new features. Thus, it can improve performance in long-term incremental learning.

In conclusion, our main contributions are summarized as follows:
\begin{itemize} 
\item A random auxiliary class augmentation method is proposed to allow the model to learn more diverse and transferable features to facilitate learning across different tasks while mimicking the incremental learning paradigm in the initial phase to counteract real forgetting.
\item We propose mixing features extracted from the new and old feature extractors to replace new features to reduce the angle between the current feature and the old one so that a less biased feature updates the model to relieve catastrophic forgetting.
\item Extensive experiments show that our method significantly outperforms SOTA NECIL methods and obtains comparable results to recent exemplar-based methods.
\end{itemize}

\section{Related Work}
\subsection{Class-Incremental Learning}
Generally, class incremental learning methods can be divided into three categories. Regularization-based methods \cite{ewc, si} calculate parameter importance in different ways to prevent parameters important for old tasks from being modified to maintain stability. \cite{lwf,lwm} adopt different distillations to retain the discriminative features of old tasks. Rehearsal-based methods \cite{zhu2023continual, replayTCSVT,luo2023class,2023ICLR} store several old class exemplars, which are then fed to the model with new coming samples. EEIL\cite{EEIL} proposes fine tuning the network by class-balanced data. PODNet\cite{PODNET} defines a new Pooled Output Distillation to maintain the stability of old knowledge. BiC \cite{Bic} mitigates model bias in favor of new classes by applying bias correction to the last fully connected layer. Structure-based methods continually expand the model's structure when learning new tasks. Foster \cite{foster} freezes the old model and creates a new module to fit new tasks, then removes insignificant dimensions and parameters to keep the model compact. \cite{grow} selects and expands important sub-network involved in the optimization process of incremental tasks. PackNet \cite{packnet} iteratively prunes the network and leaves room for new tasks. DER \cite{der} trains a new feature extractor for each new task, then merges all previous feature extractors as the current feature extractor, where a mask layer is used to reduce the feature dimensions. DNE \cite{hu2023dense} proposes a dense network method to expand the transformer by creating a new branch to accommodate each new task. Currently, relay-based methods obtain the highest performance.

Considering the storage requirement and privacy concern, the NECIL \cite{pass,ssre} has received increasing attention. \cite{ssre} introduces a residual adapter to learn new tasks, which is then incorporated into the model by re-parameterizing the model structure. Self-supervised learning and prototype augmentation are adopted in \cite{pass} to maintain the stability of the decision boundary. To learn more transferable features, IL2A \cite{il2a} introduces class augmentation and uses semanAug to generate information about old classes. R-DFCIL \cite{R-DFCIL} maintains the model's stability by distilling the angle difference between three samples. Fetril \cite{Fetril} computes the relationship between the new and old class means, then selects the old class means that are similar to the new class means for transformation as the pseudo-feature vector. Compared with these NECIL methods, we propose random class augmentation to learn sufficient transferable features and replace the new feature with mixed features to maintain old knowledge.

\subsection{Data Augmentation}
Data augmentations can effectively improve the robustness of deep neural networks. Typically, they disturb with, transform \cite{aug} or synthesize images \cite{TMMImageSynthesis, TMMGAN}. For example, Mixup \cite{mixup} performs augmentation by mixing two images and their labels. Cutout \cite{cutout} randomly cuts out a part of the image, which can make the model focus on learning global features rather than specific important regions. CutMix replaces a part of an image with a part of another image to improve the learning ability of models. In addition to these augmentations in the original image domain, some works explore semantic enhancement by transforming the representations embedded in deep feature space. The intuition is to transform the features in a particular direction corresponding to the relevant semantic augmentation. Upchurch et al. \cite{DFI} perform semantic augmentation by conducting feature interpolation. ISDA \cite{ISDA} computes the intra-class covariance matrix and samples from this matrix to perform random semantic augmentation. 

Some recent works \cite{il2a, pass} have introduced class augmentation to NECIL. IL2A \cite{il2a} expands the class by blending two images while PASS \cite{pass} performs class augmentation by rotating original images. Experiments have demonstrated the superiority of these methods. However, these methods have their drawbacks (e.g., fewer transferable features or too many mixed features). Therefore, we further propose random auxiliary class augmentation to address these deficiencies. Meanwhile, we introduce randomness to NECIL, which allows the model to mimic the incremental learning process. Therefore, it can resist forgetting better when encountering new tasks.
\begin{figure*}[t]
	\centering	\includegraphics[width=1.0\textwidth,height=0.35\textwidth]{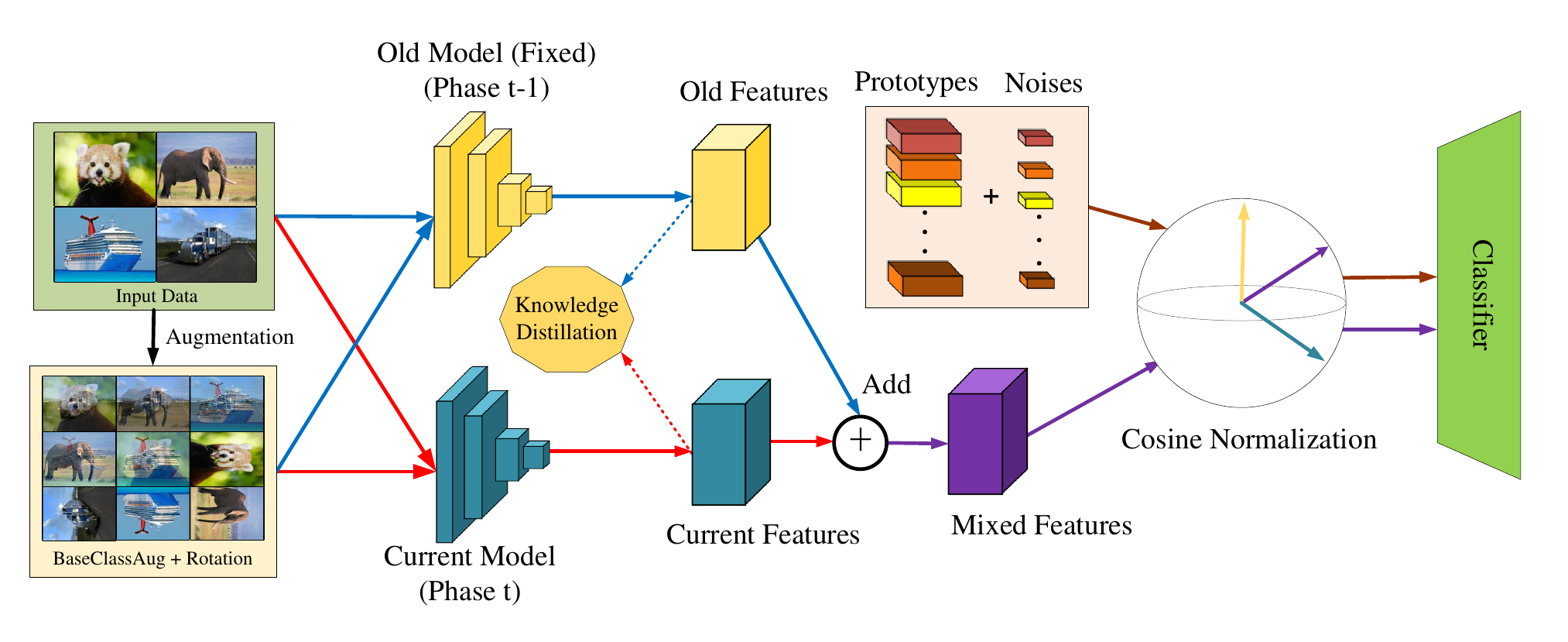}
	\caption{The RAMF framework for incremental learning phases. Fixed rotation auxiliary classes are adopted in these phases rather than the Random auxiliary classes in the initial phase.}
	\label{MIX}
\end{figure*}

\section{Methodology}
We first introduce the problem formulation of NECIL and give an overview of RAMF. Next, we present some data augmentation methods relevant to our random auxiliary class augmentation. Finally, the key components of RAMF are detailed. 

\subsection{Problem Formulation} 
In NECIL, the training streaming dataset $\mathcal{D}=\{D_{1},\ldots, D_{T}\}$ is divided into $T$ parts. Each part consists of $N_{t}$ sample pairs $(x_t,y_t)$, where $x_t\in \mathcal{X}_{t}$ and $y_t \in \mathcal{Y}_{t}$ denote the training images and corresponding labels at stage $t$. $\mathcal{X}_{t}$ and $\mathcal{Y}_{t}$ is the input space and label space respectively. At different stages, the label spaces are disjoint $\mathcal{Y}_{i} \cap \mathcal{Y}_{j} = \emptyset, i \neq j$. Note that samples of old classes can not be stored in NECIL. Hence, at the incremental stage $t$, the model is updated only using current stage data $D_{t}$ while tested on all the classes that have been learned $\mathcal{Y}_{1:t}=\mathcal{Y}_{1} \cup \mathcal{Y}_{2}   \cup ...  \cup \mathcal{Y}_{t}$. The goal of incremental learning is to train a single feature extractor $F_{\theta}$ and a classifier $G_{\phi}$, which usually is the last fully-connected layer.

\subsection{Overview of RAMF}
The RAMF consists of initial stage and incremental phases. In the former, each input image is augmented twice as shown in Figure \ref{Random}. Specifically, Base ClassAug is applied to the input batches. Meanwhile, one of three auxiliary class augmentations is selected randomly and conducted, where some augmented class labels are generated. In other words, additional nodes are added in the last fully connected layer. We compute classification loss for the augmented classes and original classes. After this initial training is completed, we remove these additional added nodes and compute the class prototypes for all original classes. In each incremental phase, the original images are also augmented twice. However, the auxiliary class augmentation is fixed to rotation. To reduce forgetting, we proposed mixed features to replace new features for learning new tasks. And we also use the model from the previous phase as a teacher model to conduct knowledge distillation. Algorithm \ref{algorithm1} illustrates the whole procedure of our RAMF framework. 

\subsection{Preliminaries} 
Before detailed the components of RAMF, we first retrospect some related data augmentation methods.
\subsubsection{Mixup}
Mixup \cite{mixup} was first proposed as a way for data augmentation in supervised learning. It can be denoted as follows:
\begin{equation}
    \tilde{x}_{mixup}=\lambda x_{A}+(1-\lambda) x_{B},
\end{equation}
\begin{equation}
    \tilde{y}_{mixup} = \lambda y_{A}+(1-\lambda) y_{B},
\end{equation}
where $x_{A},x_{B}$ are two training samples, whose label are $y_{A},y_{B}$. $\tilde{x}_{mixup}$ and $\tilde{y}_{mixup}$ are the generated sample and label. Inspired by this, IL2A \cite{il2a} proposed the classAug, which mixes samples in the same way. However, it differs from Mixup in setting the value of $\lambda$ and the label of generated samples. Specifically, classAug regards newly generated samples as extra classes. Thus, classAug expands the original $K$-classes data into $K(K-1)/2$ classes. It effectively improves the model's generalizability by allowing the model to learn more transferable features.
\begin{figure*}[t]
	\centering	\includegraphics[width=0.9\textwidth,height=0.75\textwidth]{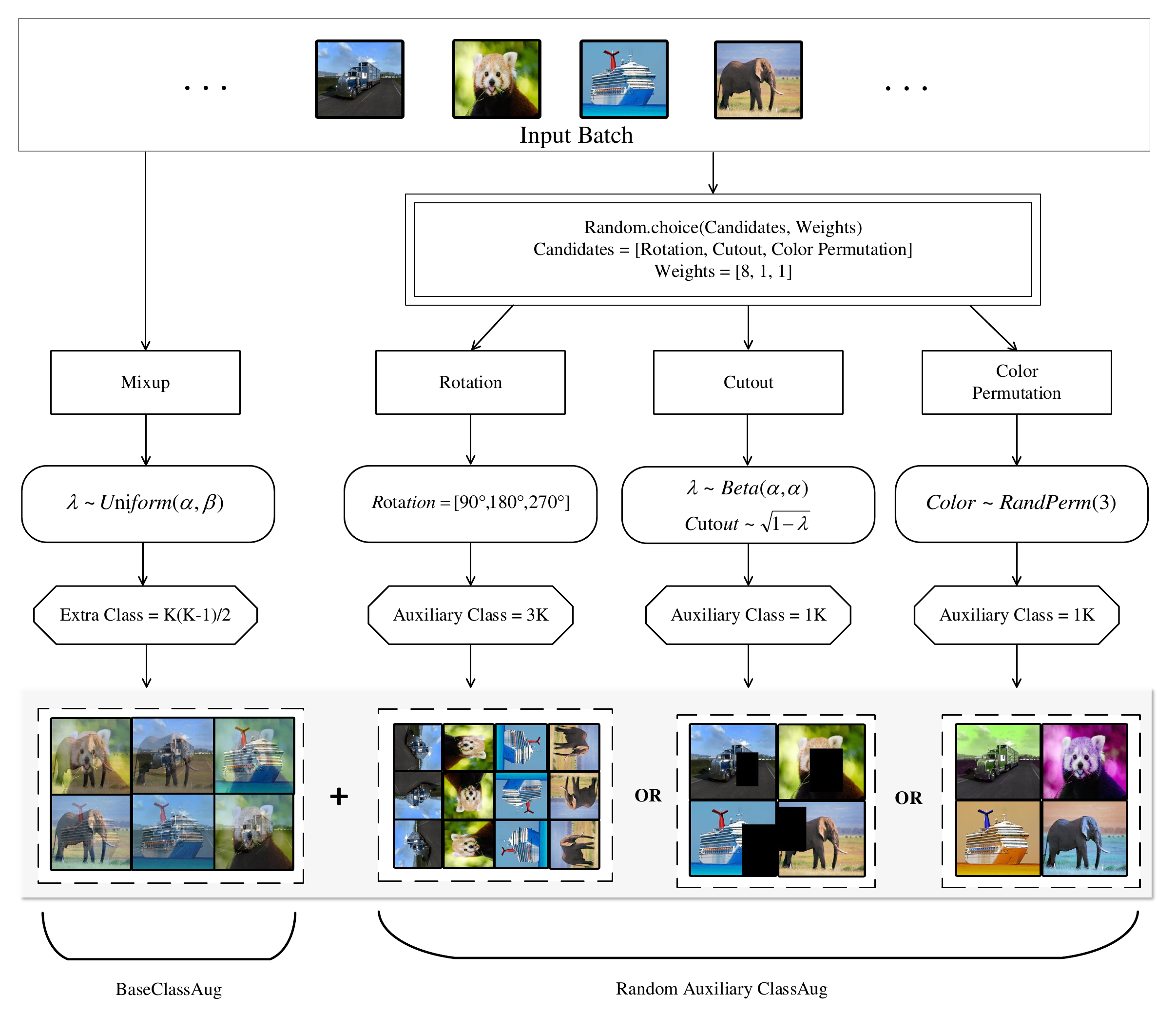}
	\caption{Class augmentation in the initial stage. For each input batch, we apply Base ClassAug and Random Auxiliary ClassAug, which randomly select one of the auxiliary class augmentations. Finally, the augmented and original images are fed into the model for learning.}
	\label{Random}
\end{figure*}

\subsubsection{Cutout} Cutout \cite{cutout} performs data augmentation by randomly cutting out a rectangular region of the sample and then filling it with 0 values. It forces the model to focus more on the global image area rather than a set of specific visual features, which can improve the model's robustness. The cutout operation can be expressed as follows:
\begin{equation}
	\tilde{x}_{cutout}=0 \times x_{M}+ x_{(1-M)},
\end{equation}
where $x_{M}$ and $x_{1-M}$ indicates the cutout area and remaining image area respectively. In the original work, the augmented data is set to the same class as the original sample. In this paper, the augmented data $\tilde{x}$ is set to a new class, thus the $K$ classes are extended to $2K$ classes through Cutout operation, which enables the model to learn more transferable features.

\subsubsection{Rotation and Color Permutation} Lee et al. \cite{colorpermutation} introduced image rotation and color permutation for label augmentation in self-supervised learning. The rotation rotates the original samples by 90°, 180°, and 270° to obtain three images with different classes. Color permutation transforms the channel RBGs into five different permutations and acquires five additional classes. The specific operations are as follows:
\begin{equation}
	\tilde{x}_{rotate}= \operatorname{rotate}\{x, \theta\}, \theta \in\{90,180,270\},
\end{equation}
\begin{equation}
	\tilde{x}_{RGB}= \{{x}_{RBG},{x}_{GRB},{x}_{GBR},{x}_{BRG},{x}_{BGR}\},
\end{equation}
where $\tilde{x}_{rotate}$ and $\tilde{x}_{RGB}$ denote the augmented samples derived from rotation and color permutation, respectively. If we directly adopt these settings, the high similarity between different classes brought by color perturbation may lead to model over-fitting, which affects transferability. Therefore, we set the different color permutations to the same extra class to adapt to continuous learning scenarios. In a word, for $K$ classes, the rotation can produce an additional $3K$ classes, while color permutation generates different $K$ classes.

\subsection{Random Auxiliary Classes Augmentation}\label{Auxiliary Classes}
In incremental learning, preserving old knowledge means keeping previously learned feature embedding stable. Through spectral decomposition, Zhu et al. \cite{il2a} analyzed which part of feature representations tends to be forgotten and may not be transferable across different tasks. They first decomposed and denoted the feature representation as a group of eigenvectors. Then, they proposed to employ the cosine value of the angle between eigenvectors to measure the representation shift between old and updated feature extractor during incremental learning. Based on this metric, they draw a conclusion through experiments, i.e. the directions with larger eigenvalues transfer better and suffer less forgetting. And a representation should have the following properties: (1) the eigenvalues of the features should be enlarged to transfer across tasks (i.e. larger transferability) (2) the number of directions with significant eigenvalues should be increased. 

Based on the above analysis, we propose an effective random class augmentation method that provides sufficient extra classes to allow the model to learn substantial transferable features. To obtain more diverse samples and extra classes, we perform both Base ClassAug and Auxiliary ClassAug on each input batch. We always use the mixup as the Base ClassAug while the latter is different in the initial and incremental stages. To enable the model to learn more transferable and diverse features, we use random auxiliary class augmentation in the initial phase. Specifically, we randomly select one augmentation method from three candidates as shown in Figure \ref{Random}. 
This selection process can be expressed in the following manner:
\begin{equation}
	\textit{random.choices}(\mathcal{C}\textit{andidates}, \mathcal{W}\textit{eights}),
\end{equation}
where $\mathcal{C}\textit{andidates}$ and $\mathcal{W}\textit{eights}$ are augmentation candidates and their weights. Note that the $\mathcal{C}\textit{andidates}$ have a one-to-one correspondence with $\mathcal{W}\textit{eights}$. In this paper, we set the $\mathcal{C}\textit{andidates}$ as $[\mathit{Rotation, Cutout, Color Permutation}]$ and $\mathcal{W}\textit{eights}$ as $[8,1,1]$, whose effect will be discussed in the experimental section. Through BaseClassAug and random auxiliary ClassAug, the label space of our method is further expanded to $K(K-1)/2 +\sum_{i=1}^{i=n}A_i$, where $A_i$ is the number of additional labels from each auxiliary class augmentation.

In the following incremental learning phases, we fix Auxiliary ClassAug as the rotation augmentation for two reasons. The first is that compared with other augmentation methods, the rotation can produce more features and classes from original images, which means more transferable features are available. The second is that the appearance of the random classes will update the model a big step on another optimization path. However, the model must minimize the feature space drift in the incremental phase to maintain stability. Therefore, fixed Auxiliary ClassAug can avoid excessive swings of fine-tuned parameters, which is beneficial to keep the previously learned feature space. Note that setting the color permutation to $7K$ extra classes yields more extra classes than rotation. However, this can trap the model to task-level overfitting and detrimental to incremental learning.

\textbf{Discussion.} Compared to the previous methods \cite{il2a, pass}, our random auxiliary class augmentation method offers several advantages. Firstly, it encompasses a larger number of extra classes, further enlarging the eigenvalues of the features, which enables the model to learn more transferable features and perform better in continuous learning scenarios. Secondly, it reduces the difficulty of differentiating between original and augmented images since the auxiliary augmentation contains all the original features and shares a similar feature space with the original images. 

The proposed Random Auxiliary ClassAug is a little related to RandomMix \cite{RandomMix}, which randomly selects one image mixing method from different candidates to perform data augmentation. However, the methods in RandomMix are all based on image blending. Moreover, the new labels are a combination of original labels, which has limited effectiveness in incremental learning scenarios. Since the number of classes has stayed the same, the model's transferability changed a little. In contrast, our random auxiliary class augmentation divides the augmentation into Base ClassAug and Auxiliary ClassAug. Base ClassAug is invariably performed in each input batch to make the model learn more features. Then one of the different auxiliary class augmentations is selected randomly. Furthermore, our method expands the label space to learn more transferable to boost the effect of data augmentation for incremental learning.

\begin{figure}[t]
	\vspace{-0.3cm}
	\centering  
	\subfloat[eigenvalues of representations]{
        \hspace{-0.2cm}
		\includegraphics[scale=0.30]{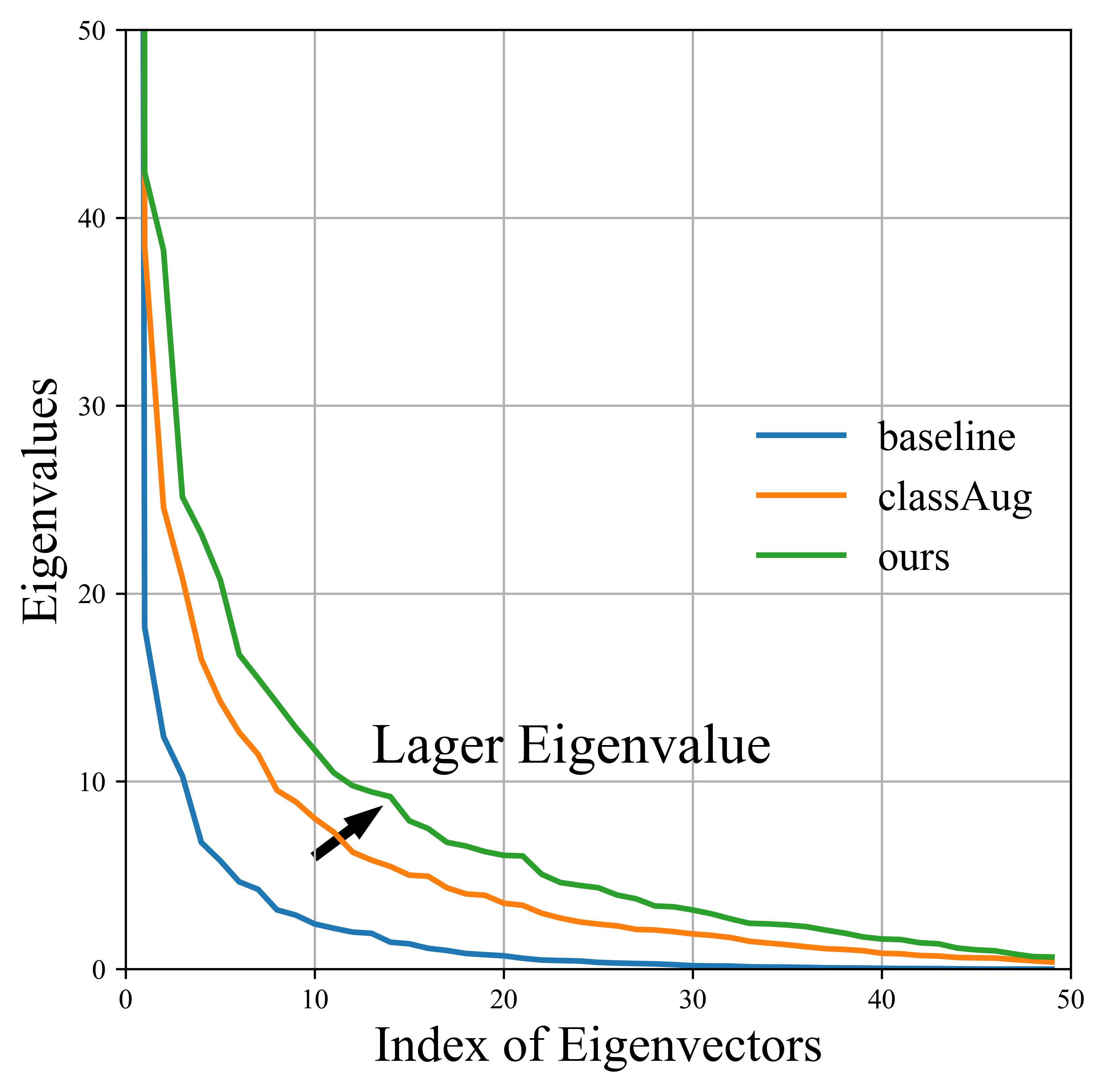}}\subfloat[cosine values of corresponding angles]{
		\includegraphics[scale=0.30]{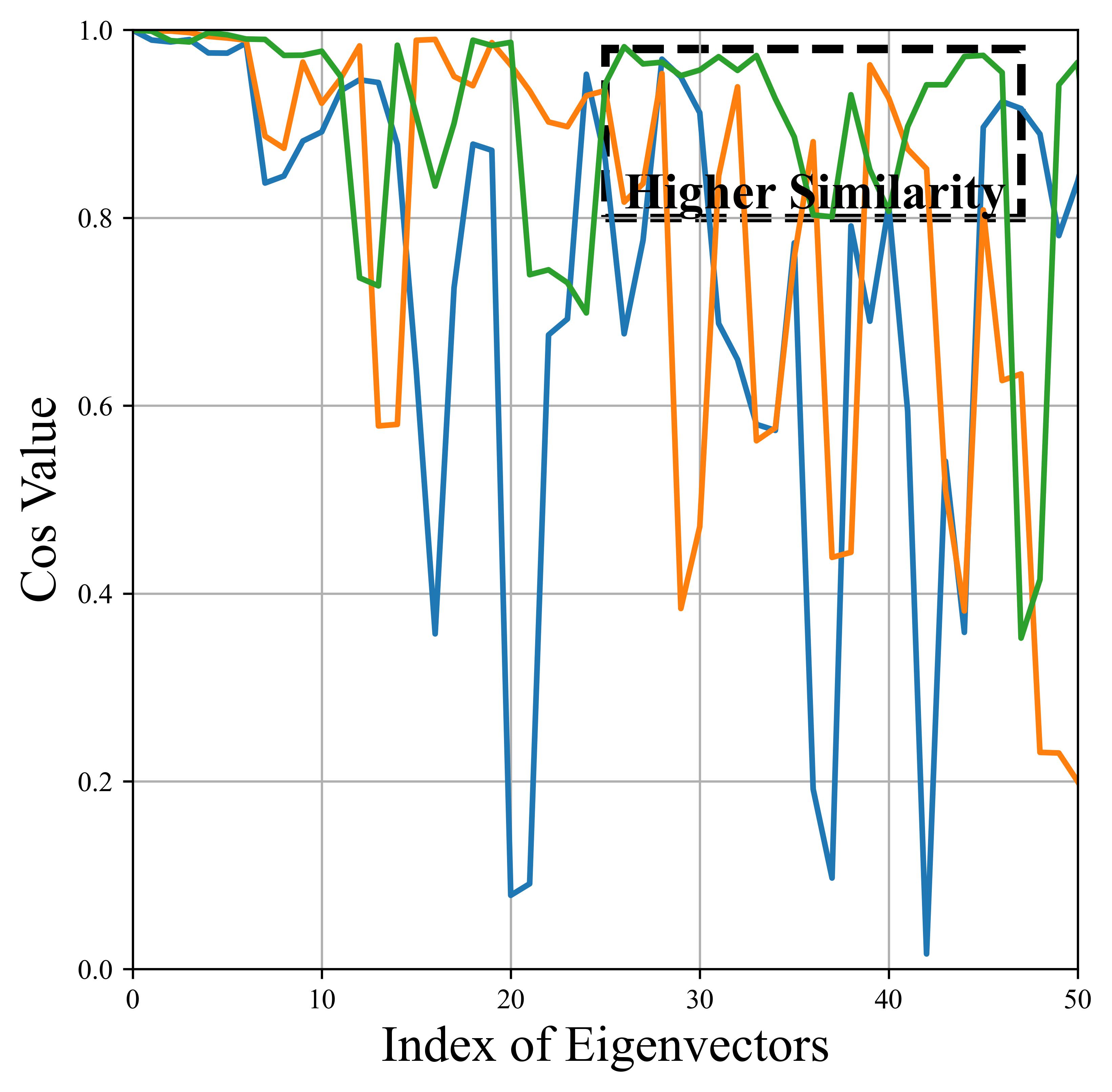}}\\
	\caption{(a) Distribution of eigenvalues for baseline, classAug \cite{il2a} and our Random Auxiliary classAug. (b) Absolute cosine values of corresponding angles.}
	\label{eigenvalue}
\end{figure}

We further compare the eigenvalue and cosine value of the angles to illustrate the advantage of our method. We use the Noisy prototype \ref{Noise Prototype} and Knowledge Distillation \ref{KD} as the baseline, then add the classAug \cite{il2a}, and finally employ our proposed Random Auxiliary Classes Augmentation. These models using ResNet-18 \cite{resnet} were trained on CIFAR-100 \cite{cifar} dataset, with 50 classes learned for initial stage and the remaining 50 classes trained subsequently in one step. It can be observed from Figure \ref{eigenvalue} that our features with larger eigenvalues have higher similarity in more directions (small corresponding angle). This indicates that our model can learn more transferable features. 

\subsection{Mixed Feature} \label{MF}
In incremental step $t$, only the samples for new classes $D_t$ are available for model optimization. In this case, learning a substantial number of new features simultaneously with a few old class representations will result in a biased model towards new classes. To reduce this bias and keep the feature space stable when learning new tasks, we propose to replace the new class features with mixed features. Specifically, if we define $f_{t}$, $f_{t-1}$ as features extracted by the current feature extractor $F_{t}$ and  old feature extractor $F_{t-1}$ respectively, the mixed features can be calculated as follow:
\begin{equation} 
	f_{mix} = \lambda * f_{t} + (1-\lambda) * f_{t-1}
\label{eq6}
\end{equation} 
where $0 \leqslant \lambda \leqslant 1 $ is the coefficient controlling the weights of different features.

\begin{figure}[t]
	\centering	\includegraphics[width=0.45\textwidth,height=0.25\textwidth]{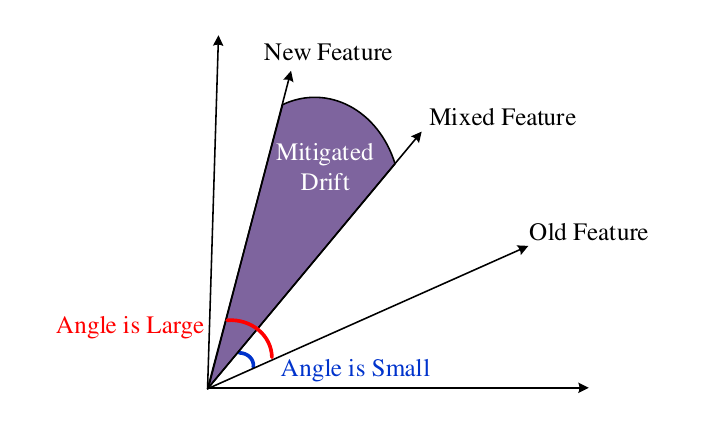}
	\caption{Explanation of how mixed features mitigate feature drift.}
    \label{fig1}
\end{figure}

As illustrated in Fig \ref{fig1}, because the features are generally greater than zero (determined by Relu), the angle between mixed feature and old feature is generally smaller than that between new feature and old feature, which also means greater cosine similarity. Formally, we can draw this conclusion through the following proof.

\noindent \textbf{Theorem 1: $cos(f_{t-1}, f_{mix}) \geqslant cos(f_{t-1}, f_t)$}
\begin{flalign*} 
\begin{split}
&\  
Proof :\cos \left(f_{t-1}, \lambda f_{t}+  \mu f_{t-1}\right)-\cos \left(f_{t-1}, f_{t}\right) \\
&=\frac{\lambda f_{t-1} f_{t}+\mu \left| f_{t-1}\right|^{2}}{\left|f_{t-1}\right|\left | \lambda f_{t}+\mu f_{t-1} \right|}-\frac{f_{t-1} f_{t}}{\left|f_{t-1}\right|\left|f_{t}\right|} \\
&=\frac{1}{A}\left[\lambda f_{t-1} f_{t}  \left|f_{t}\right|+ \mu | f_{t-1}|^{2}  \left| f_{t}\right|-f_{t-1}  f_{t}\left|\lambda f_{t}+ \mu f_{t-1}\right| \right] \\
&\geqslant \frac{1}{A}\left[\mu \left| f_{t-1}\right|^{2} \left|f_{t}\right|- \mu f_{t-1} f_{t} |f_{t-1}| \right] \\
&=\frac{\mu| f_{t-1} \mid}{A}\left[\left|f_{t-1}\right|\left|f_{t}\right|-f_{t-1} f_{t}\right] \\
&=\frac{\mu\left|f_{t-1}\right|^{2}\left|f_{t}\right|}{A}[1-\cos \theta] \geqslant 0
\end{split}&
\end{flalign*}
where $A = |f_{t-1}||f_t||\lambda f_t + \mu f_{t-1}|$ and $\mu = 1-\lambda$.
The above theorem also indicates a large similarity between mixed features and old features. Thus, the mixed features can mitigate the drift by narrowing the angle with the old features. By assigning different weights to the coefficients, the mixed features can be adapted to different incremental learning scenarios, which will be discussed in the experimental section.

According to the classical neural network architecture, the predicted probability for the mixed feature $f_{mix}$ by a classifier will be calculated:
\begin{equation}
	p_{i}(x)=\frac{\exp \left(\theta_{i}^{\mathrm{T}} f_{mix}(x)+b_{i}\right)}{\sum_{j} \exp \left(\theta_{j}^{\mathrm{T}} f_{mix}(x)+b_{j}\right)}
\end{equation}
where $p_{i}(x)$ is the predicted probability of a sample $x$, $\theta$ and $b$ are the parameters of classifier. However, adding new and old features in incremental phases may produce a larger feature value than the initial phase, which may bring prediction errors. Therefore, to decrease the negative effect of the addition in mixed feature, we follow \cite{lucir} to employ cosine normalization to replace the multiplication operation in the last layer. The probability calculation using cosine normalization is as follows:
\begin{equation}
	p_{i}(x)=\frac{\exp \left(\eta\left\langle\bar{\theta}_{i}, \bar{f_{mix}}(x)\right\rangle\right)}{\sum_{j} \exp \left(\eta\left\langle\bar{\theta}_{j}, \bar{f_{mix}}(x)\right\rangle\right)}
\end{equation} 
where $\bar{\theta}= \theta/\|\theta\|_{2}$ denotes the $l_2$-normalized vector, and $\left\langle\bar{\theta}_{1}, \bar{\theta}_{2}\right\rangle=\bar{\theta}_{1}^{\mathrm{T}} \bar{\theta}_{2}$ measures the cosine similarity between two normalized vectors. In this equation, a learnable parameter $\eta$ is introduced to control the peakiness of softmax distribution.

\subsection{Noisy Prototype} \label{Noise Prototype}
Since old class samples cannot be stored, most methods use prototypes to represent old class exemplars to avoid the feature space being entirely dominated by the new classes \cite{il2a,ssre}. The prototype for class $m$ is computed as follows:
\begin{equation}
	\text { Prototype }_{m}=\frac{1}{\mathcal{N}_{m}} \sum_{n=1}^{\mathcal{N}_{m}} F_{\theta}\left(X_{n}\right),
\end{equation}
where $ \mathcal{N}_{m} $ denotes the number of samples of class $m$. Further, PASS \cite{pass} introduced prototype augmentation to stabilize decision boundaries. Specifically, it adds Gaussian noise to prototypes and controls the degree of augment by a coefficient $r$. However, the calculation of this coefficient needs to store the covariance matrix of old classes, which increases memory requirement. 

To reduce this memory requirement and increase randomness, we generate the coefficient $r$ randomly. In detail, when learning new tasks, each old class prototype is augmented as follows:
\begin{equation}
	\text { Prototype }_{m,aug}= \text { Prototype }_{m} + e \times r,
\end{equation}
where $e\sim\mathcal{N}(0,1)$ is a Gaussian noise and $r\sim uniform(\alpha, \beta)$ is a coefficient. We then feed the augmented prototype and the feature of new data into the classifier simultaneously.

\subsection{Overall Objective of RAMF} \label{KD}
To optimize the model's parameters, the overall loss function of RAMF consists of three parts:
\begin{equation}
	L_{\text { RAMF }}=L_{\text { ce }}+\alpha * L_{N\operatorname{Porotype}}+\beta * L_{K D}
	\label{loss}
\end{equation}
where $L_{\text { ce }}$ and $L_{N\operatorname{Porotype}}$ denote the cross-entropy loss for new classes and noisy prototypes respectively. $L_{K D}$ is the KD loss. $\alpha, \beta$ are two hyper-parameters, which are set to 10 through experiments.

The former two loss functions are used to correctly classify the new class samples and old prototypes, which can be calculated as follows:
\begin{equation}
	L_{ce}(x)=-\sum_{i=1}^{N_{t}} y_{i} \log \left( G_\phi(F_\theta(x_i)) \right)
\end{equation}
\begin{equation}
	L_{N\operatorname{Porotype}}=-\sum_{i=1}^{M_{t-1}} y_{i} \log \left( G_\phi(\operatorname{Prototype}_{i,aug})\right)
\end{equation}
where $N_{t}$ is the number of samples at stage $t$, $M_{t-1}$ is the number of classes learned up to stage $t-1$, $x_i$ is an input image, whose ground-truth label is $y_i$.

To avoid large variations of feature space caused by the update of the feature extractor, we further use knowledge distillation \cite{KD} to regularize the feature extractor. Specifically, it restricts the drift of features extracted by the current and previous feature extractor $F_t$, $F_{t-1}$
\begin{equation}
	L_{K D}=\left\|F_{t}\left(X_{t} ; \theta_{t}\right)-F_{t-1}\left(X_{t} ; \theta_{t-1}\right)\right\|_{2}
\end{equation}

\section{Experiments}
In this section, we conduct extensive experiments to verify the effectiveness of RAMF.
\begin{algorithm}[t] \label{algorithm1}
    \SetAlgoLined 
    \textbf{Initialization:} training dataset streaming  $\mathcal{D}$\; Learning Rate $\lambda$, Parameters $\theta,\phi$\; 
	\caption{RAMF Training Algorithm}
	\ForEach{incremental step  $t \in\{1, \ldots,  T\}$}{
    \textbf{Input}: $D_{t} = (x_t,y_t)$;\\
	\textbf{Output}: $model_{t}$;\\
		\eIf{t=1}{
		 perform Base Class Augmentation;\\
		 select one Auxiliary ClassAug Randomly;\\
		 optimize $\theta^{0}$, $\phi^{0}$ by minimizing $L_{t,ce}$;\\
	     $model_{1} \gets  model_{0}$;\\
		 remove augmented class nodes in classifier;\\
		 compute and save $prototype_{D_{t}}$;\\
		}
     {
        perform Base class Augmentation;\\
		perform Rotation Class Augmentation;\\
		compute Mixed features;\\
		compute Noisy prototype;\\
		optimize $\theta^{t}$, $\phi^{t}$ by minimizing Eq.\eqref{loss};\\
		$model_{t} \gets  model_{t-1}$;\\
		remove augmented class nodes in classifier;\\
		compute and save $prototype_{D_{t}}$;\\ 
       }
       }
\end{algorithm}
\subsection{Datasets and Settings.}
\subsubsection{Datasets} Following previous works, we employ three popular benchmarks CIFAR-100 \cite{cifar}, TinyImageNet \cite{tiny} and ImageNet-Subset. Specifically, CIFAR-100 contains 60,000 images of 32$\times$32 size covering 100 classes, each of which contains 500 training images and 100 test images. TinyImageNet includes 200 classes, each with 500 training images, 50 test images and 50 validation images. ImageNet-SubSet, a subset of ImageNet \cite{ImageNet} contains 100 classes, each with 700 training images, 400 validation images and 200 test images. 

We follow previous works \cite{il2a, pass} for the division of classes in the initial phase and incremental phases. Moreover, we also add 5 and 20 phases for ImageNet-Subset to further analyze the performance of different methods. Therefore, for CIFAR-100 and ImageNet-Subset, the three divisions are (1) 50 initial classes with 5 incremental stages of 10 new classes, (2) 50 initial classes with 10 incremental stages of 5 new classes, (3) 40 initial classes with 20 incremental stages of 3 new classes. For TinyImageNet, the initial stage contains 100 classes, and the increments are (1) 5 stages of 20 classes, (2) 10 stages of 10 classes, (3) 20 stages of 5 classes. These different configurations allow for a comprehensive comparison of various methods.

\subsubsection{Evaluation Metrics} Following \cite{ssre, pass,il2a}, the \textit{average accuracy} is mainly used to evaluate the general performance. At step $t$, it is defined as 
\begin{equation}
    A_{t}=\frac{1}{t} \sum_{i=1}^{t} a_{t, i}
\end{equation}
where $a_{t,i} \in [0,1]$ represents the accuracy of task $i$ after learning $t$ tasks. 

Besides, we also use the \textit{average forgetting} and \textit{average intransigence} to further measure the stability and plasticity of a model. In phase $t$, the \textit{average forgetting} is defined as follows:
\begin{equation}
    F_{t}=\frac{1}{t-1} \sum_{i=1}^{t-1} f_{t}^{i}
\end{equation}
where $f_{t}^{i}=\max \limits_{k \in \{ 1, \ldots, t-1\} }\left(a_{t, i}-a_{k, i}\right)$ is the forgetting value of the $i$-th task after learning $t$-th task. $ F_t \in[0,1]$ indicates how much the model has forgotten about the old task. And the lower forgetting indicates a more stable model. 

The \textit{average intransigence} is defined as follows:
\begin{equation}
   I_{t}=\frac{1}{t} \sum_{i=1}^{t} a_{i}^{*}-a_{i, i},
\end{equation}
where $a^*_i$ denotes the accuracy of task $i$ by a reference model. This metric indicates the model's plasticity. And the lower $I_t$ is, the higher model plasticity.
\begin{table*}[t] 	
	\centering
	\begin{small}
		\renewcommand{\arraystretch}{1.3}
		\setlength{\tabcolsep}{3mm}{
			\caption{Comparisons of the average accuracy (\%) with other methods on CIFAR-100, TinyImageNet, and ImageNet-Subset. P represents the number of phases and E represents the number of stored exemplars.}
			\begin{tabular}{l|l|ccc|ccc|cccc}

				\hline
				\multicolumn{2}{c|}{} & \multicolumn{3}{c|}{\textbf{CIFAR-100}} & \multicolumn{3}{c|}{\textbf{TinyImageNet}} & \multicolumn{3}{c}{\textbf{ImageNet-SubNet}} \\ \cline{3-5} \cline{6-8} \cline{9-11}
				
				\multicolumn{2}{c|}{\multirow{-2}{*}{\textbf{Methods}}} & \textit{P=5} & \textit{P=10} & \textit{P=20} & \textit{P=5} & \textit{P=10} & \textit{P=20} & \textit{P=5}& \textit{P=10}& \textit{P=20} \\ 
				\hline
				&iCaRL-CNN(17'CVPR) &51.07 &48.66 &44.43 &34.64 &31.15 &27.90 &53.62 &50.53&42.19\\[0.5ex] 
				&iCaRL-NCM(17'CVPR) \cite{icarl} &58.56 &54.19 &50.51 &45.86 &43.29 &38.04 &65.04&60.79&53.41\\[0.5ex] 
				&BiC(19'CVPR) \cite{lucir} &61.47 &55.05 &52.03 &48.69 &40.18 &35.12 &68.31 &63.32&55.81\\ [0.5ex] 
				\multirow{-4}{*}{\rotatebox{90}{\textit{(1) E=20}}}
				&PODNet(20'ECCV) \cite{PODNET} &64.34 &62.63 &59.59 &56.32 &54.53 &52.40  &75.54&74.33&69.14\\ [0.5ex] 
                &FOSTER(22'ECCV) \cite{foster} &70.10 &67.94 &- &60.07 &56.83 &52.01  &80.22&-&-\\ [0.5ex] 
				\hline

				&IL2A(21'NIPS) \cite{il2a} &64.93 &59.04 &56.83 &47.33 &43.49 &39.32  &68.05&61.02&50.16\\[0.5ex] 
				&PASS(21'CVPR) \cite{pass} &63.47 &61.84 &58.09 &49.55 &47.29 &42.07 &66.69 &61.80&52.92\\[0.5ex] 
				&SSRE(22'CVPR) \cite{ssre} &65.88 &65.04 &61.70&50.39 &48.93 &48.17  &-&67.69&-\\[0.5ex] 
				&Fetril(23'WACV) \cite{Fetril} &66.31 &65.22 &61.51&54.82&53.14 &52.26 &72.25 &71.20&67.11 \\[0.5ex] 
				
				\multirow{-5}{*}{\rotatebox{90}{\textit{(2) E=0}}}	&RAMF(Ours) &\textbf{73.35} &\textbf{73.14} &\textbf{68.97} &\textbf{59.20} &\textbf{57.15} &\textbf{52.85}&\textbf{74.86} &\textbf{74.42}& \textbf{69.51}	\\ [0.5ex] 
				\hline	
		\end{tabular}
        \label{avgacc}	
  }
	\end{small}
\end{table*} 

\begin{figure*}[t] 
\centering 
\subfloat[CIFAR-100]{	
	\includegraphics[scale=0.34]{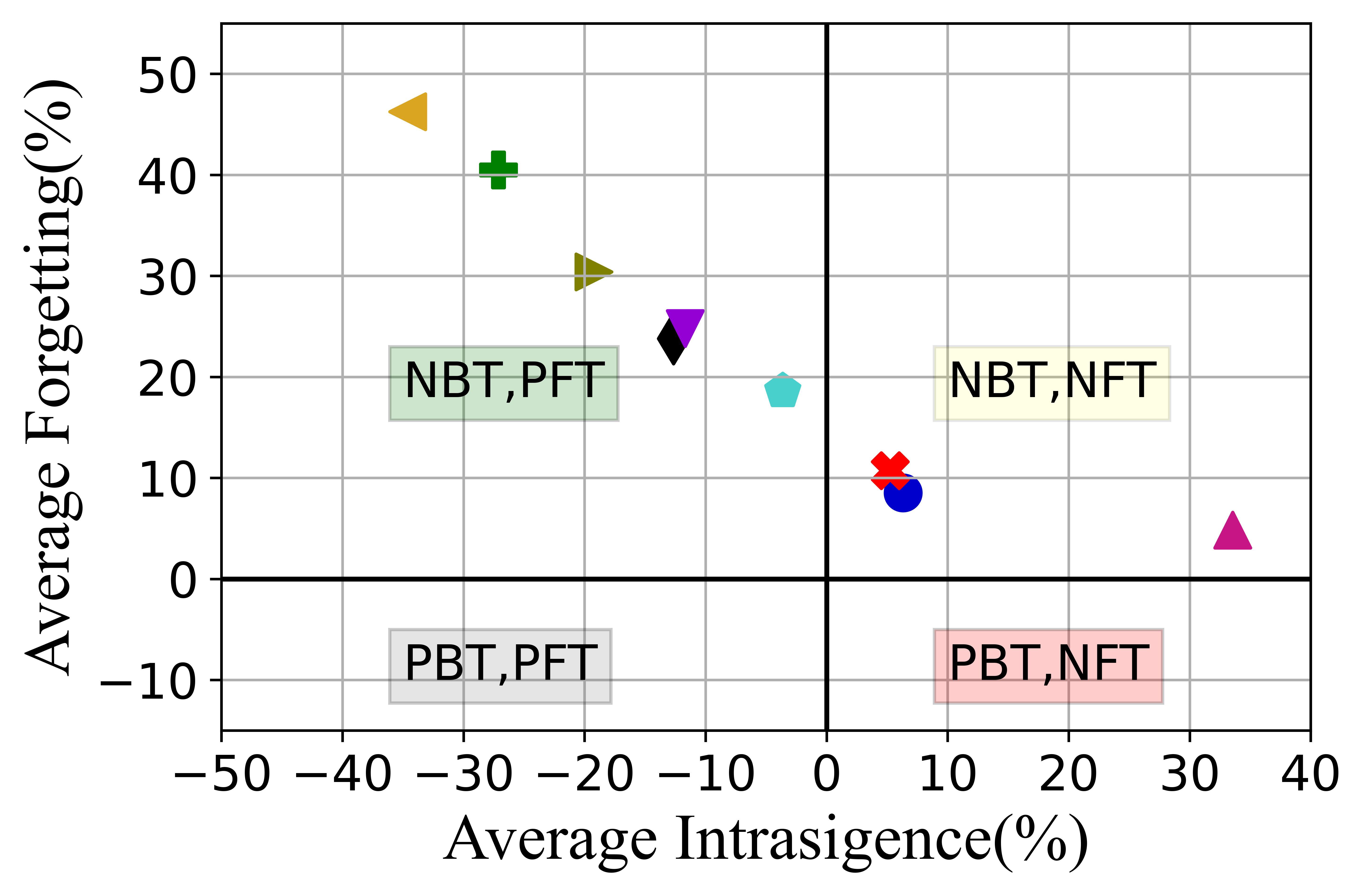}}
\subfloat[TinyImagenet]{
	\includegraphics[scale=0.34]{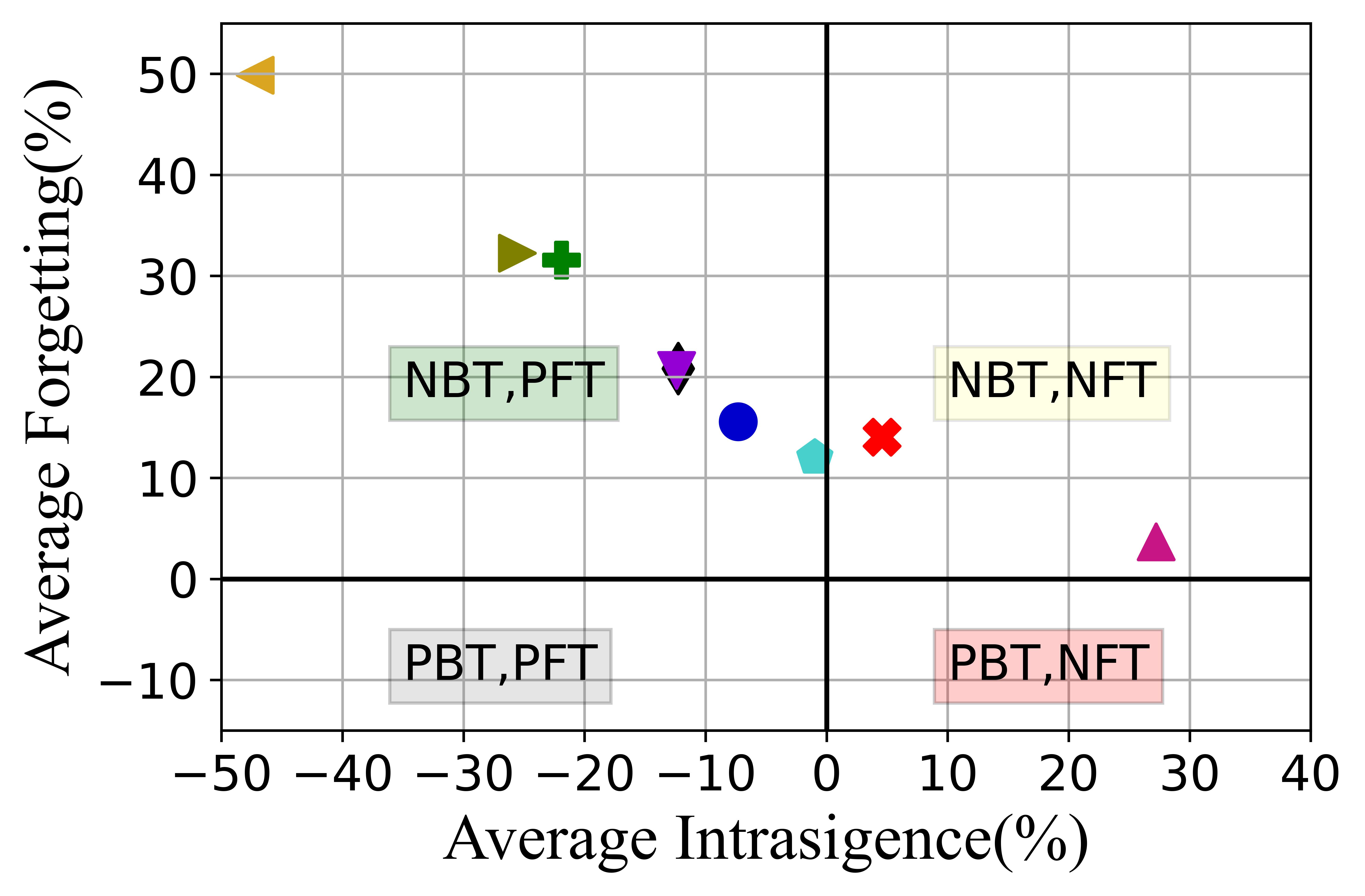}}
\subfloat[Imagenet-Subset]{
	\includegraphics[scale=0.34]{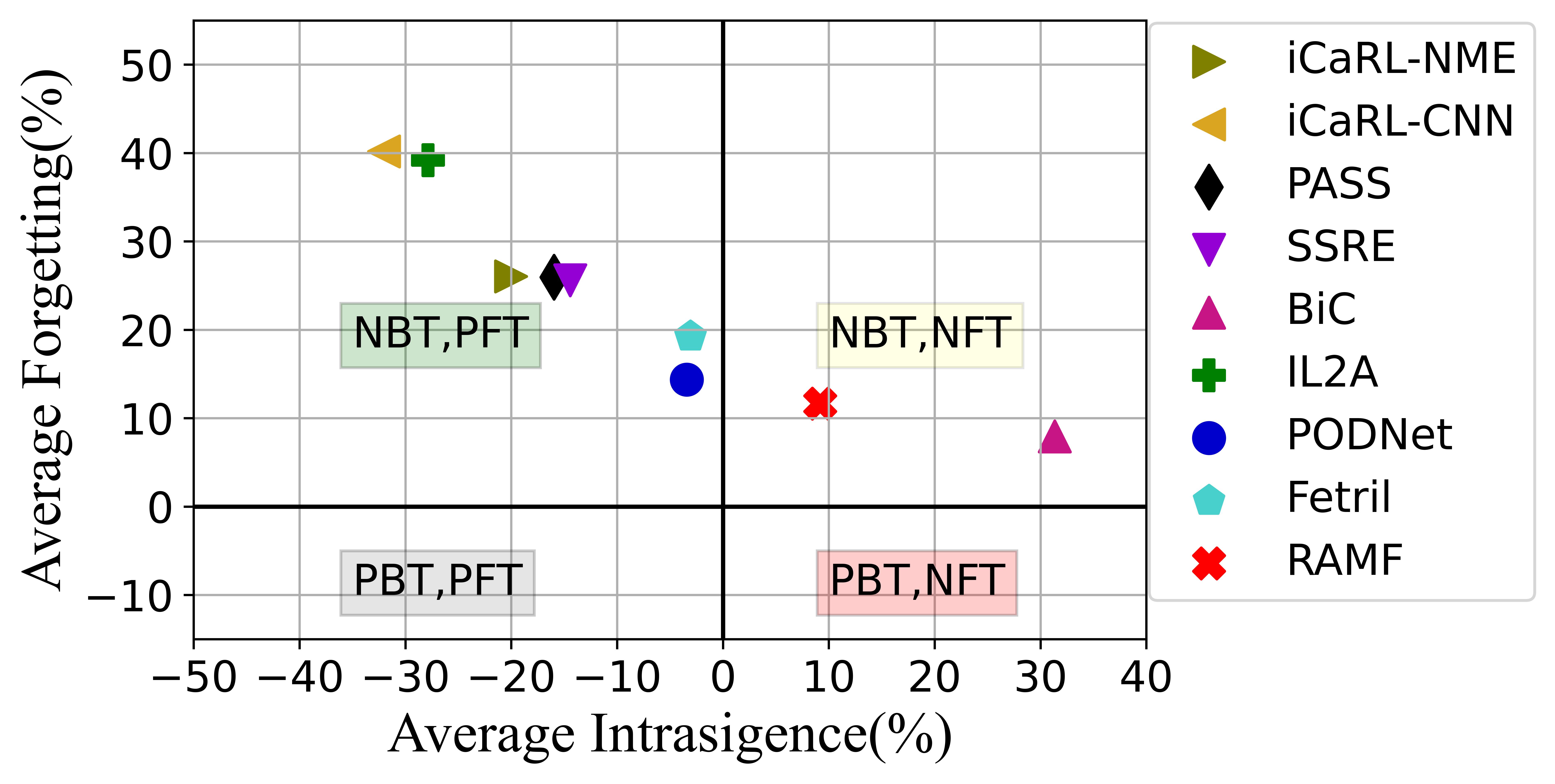}}
\caption{Interplay between average intransigence and average forgetting. PFT indicates Positive Forward Transfer, NBT indicates Negative Backward Transfer, and so on for the others.}
\label{for-intra}
\end{figure*}
\begin{figure*}[t] 
	\centering 
	\subfloat{	
		\includegraphics[width=0.31\linewidth]{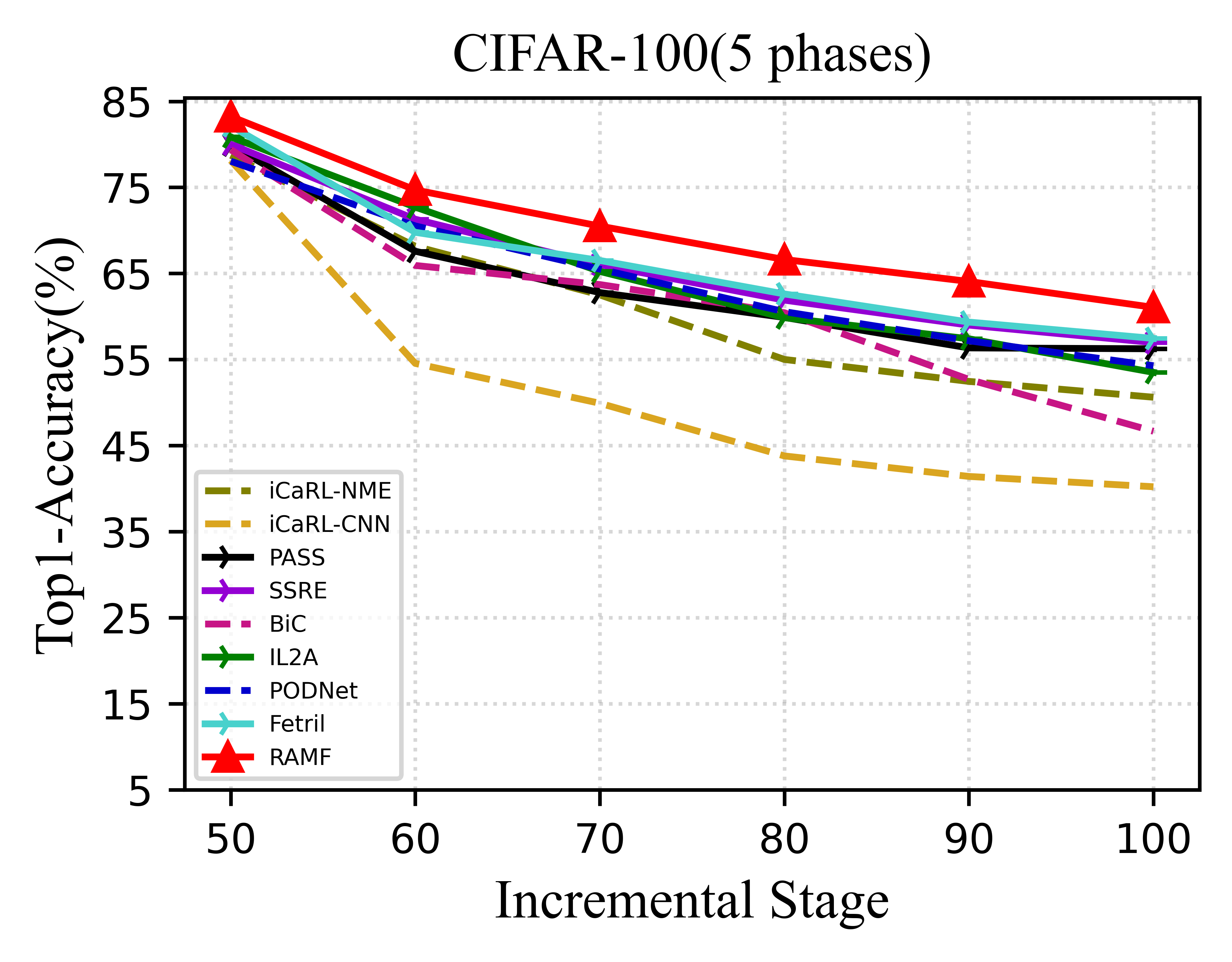}}
	\subfloat{
		\includegraphics[width=0.31\linewidth]{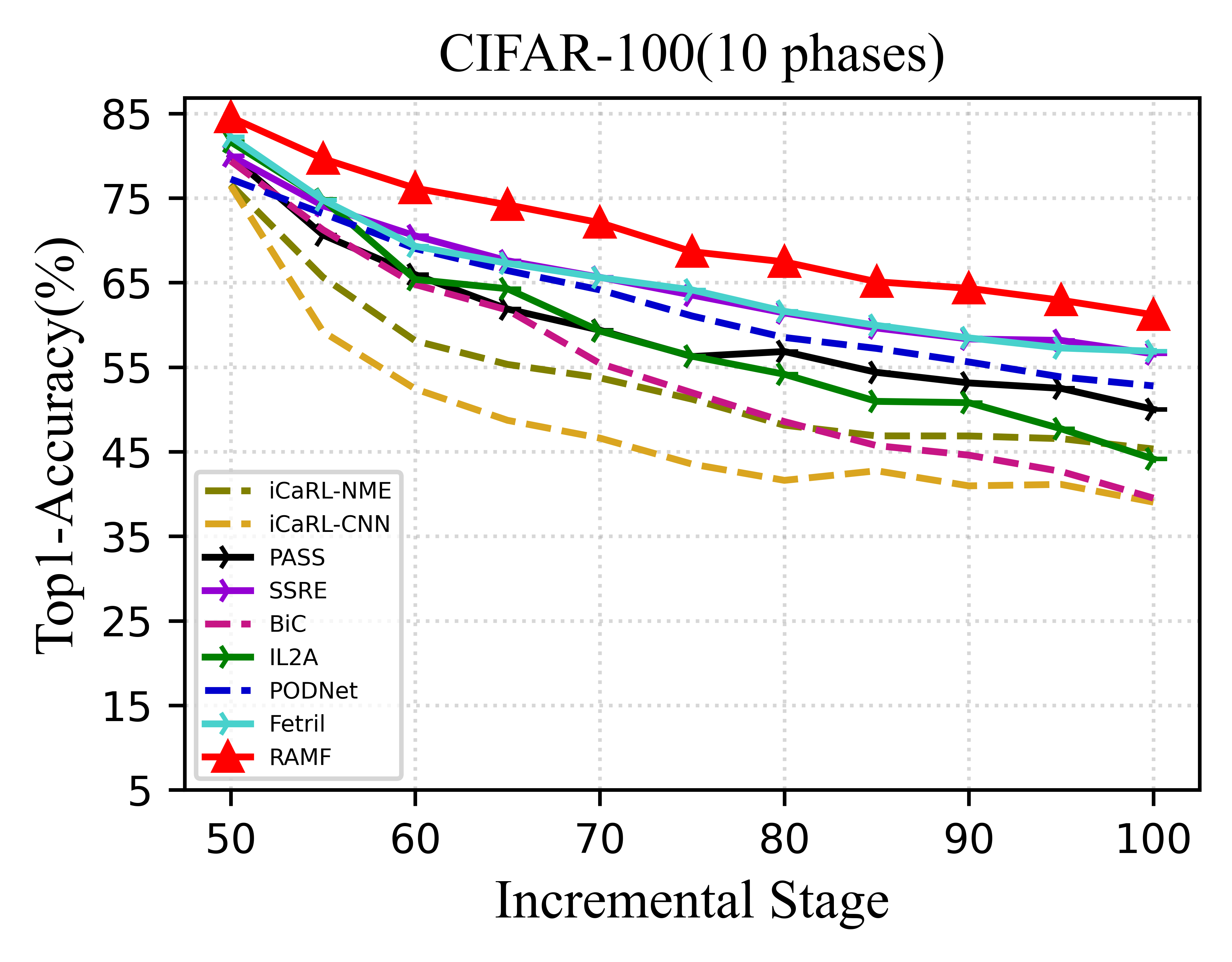}}
	\subfloat{
		\includegraphics[width=0.31\linewidth]{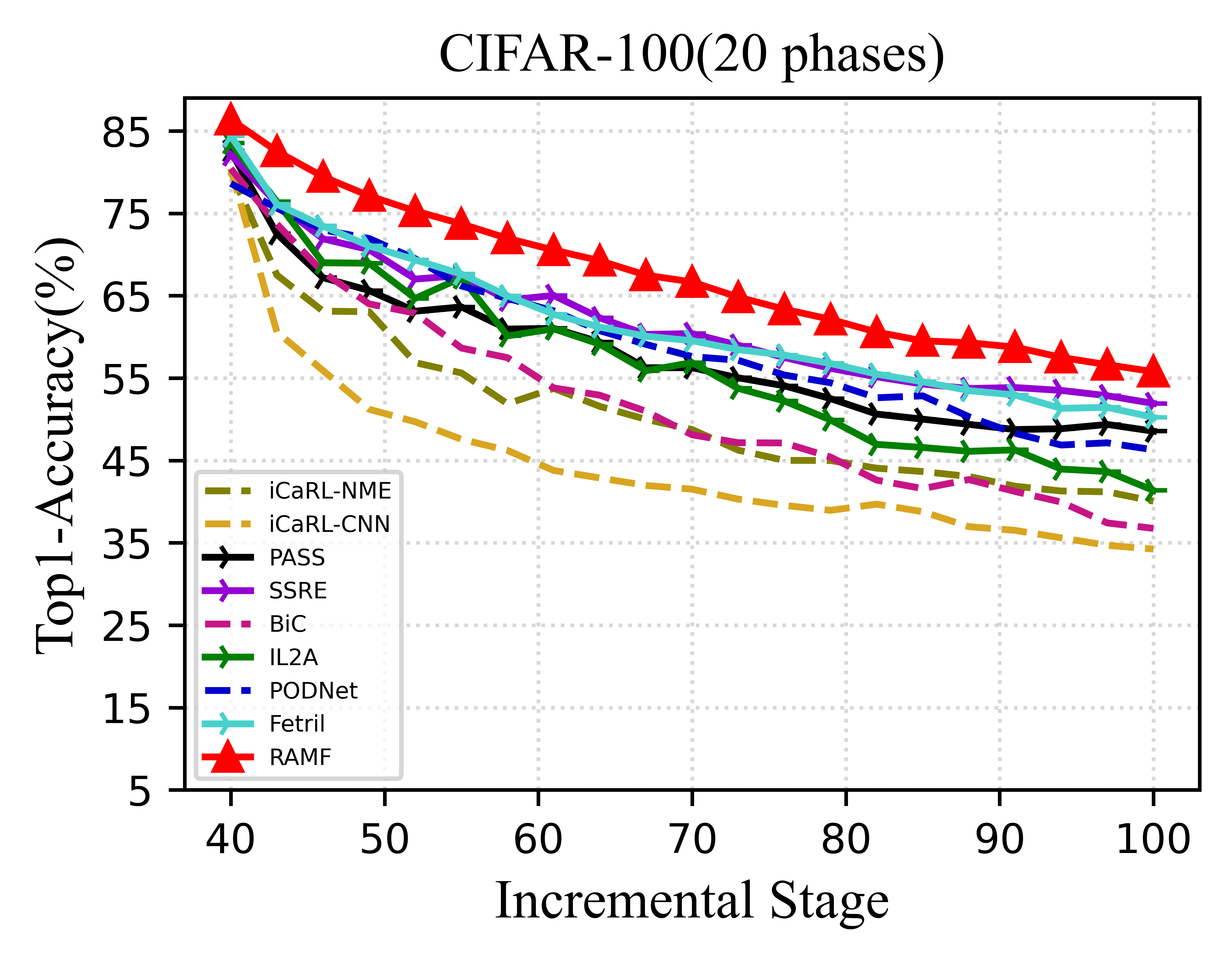}}\\
  \subfloat{	
		\includegraphics[width=0.31\linewidth]{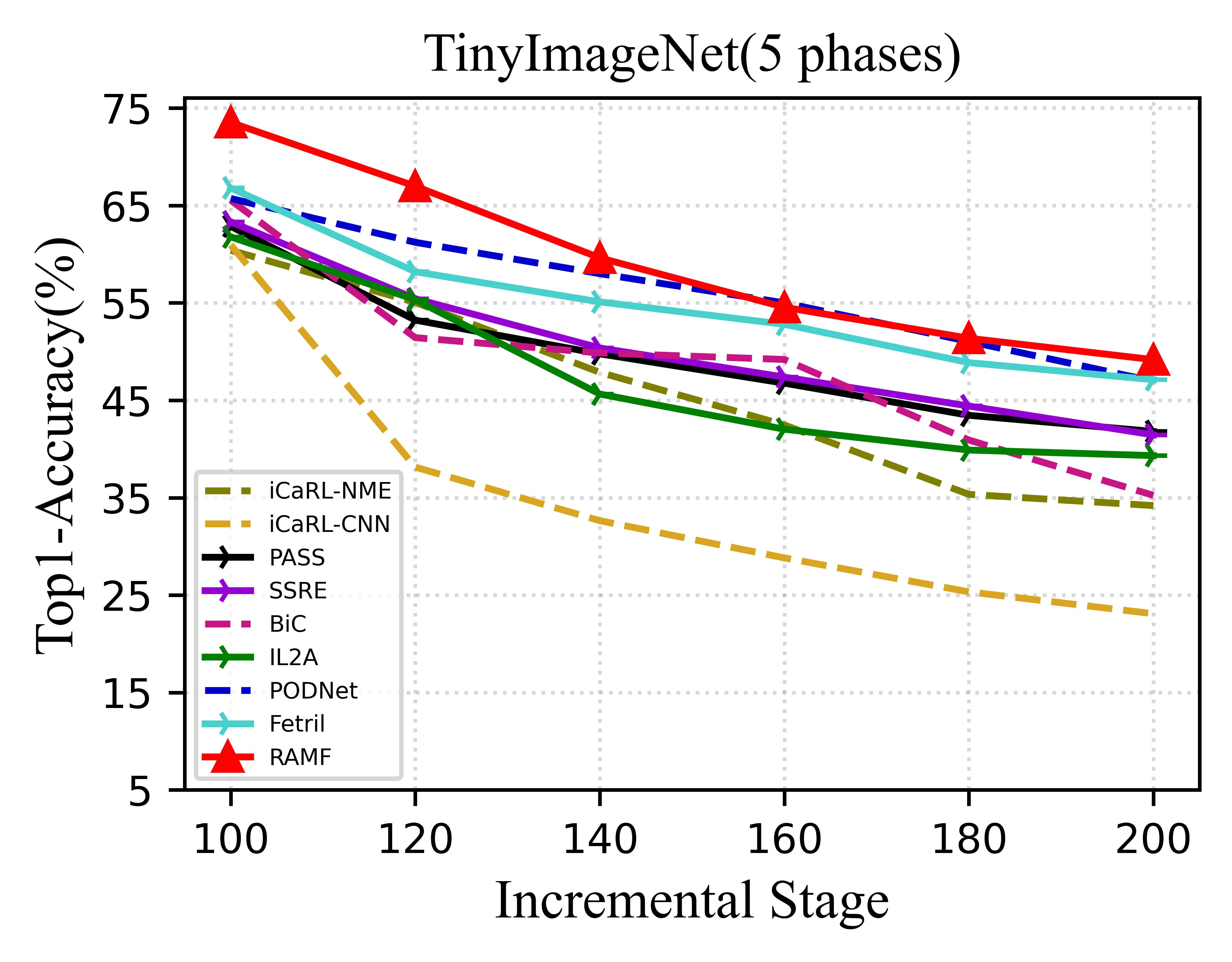}}
	\subfloat{
		\includegraphics[width=0.31\linewidth]{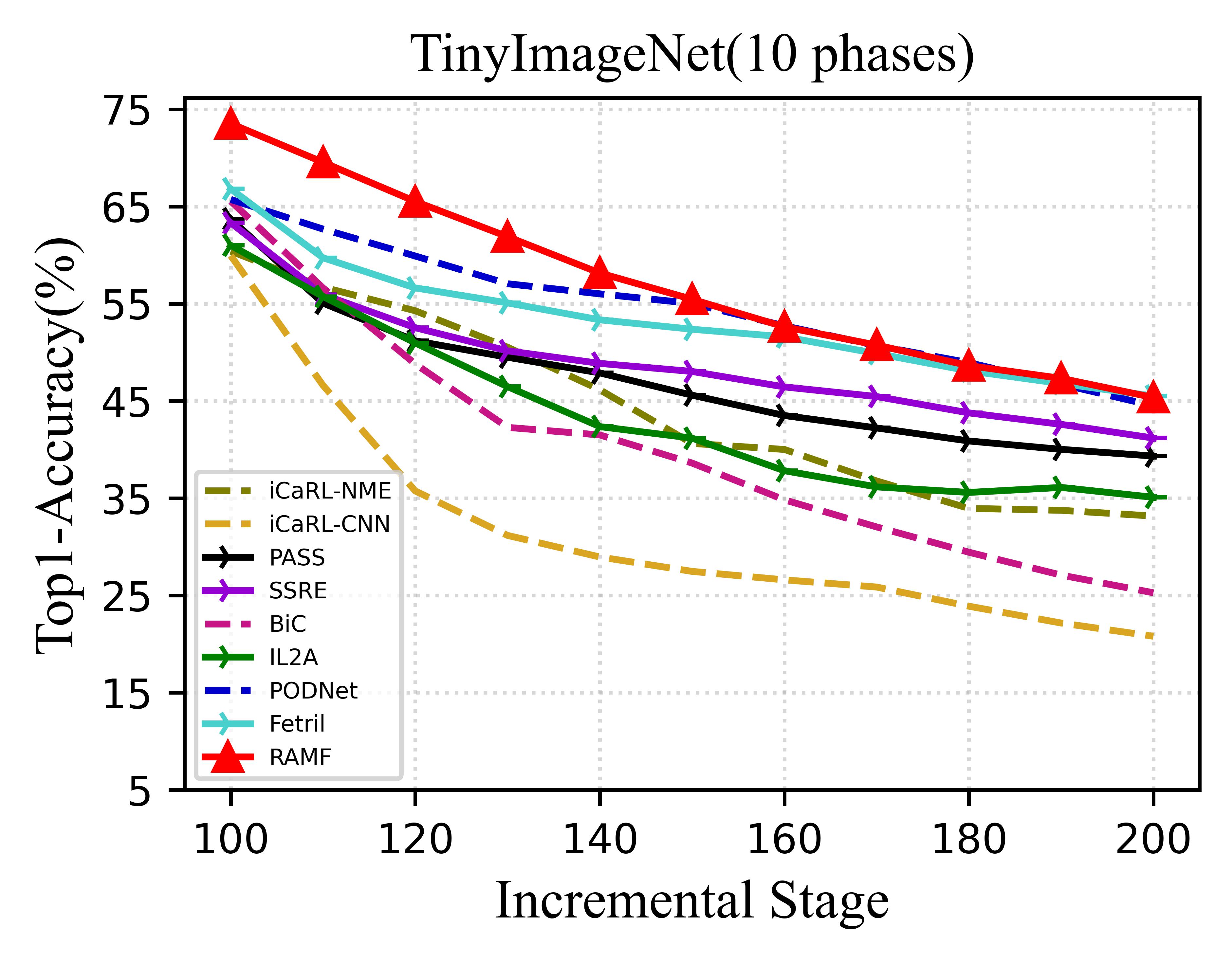}}
	\subfloat{
		\includegraphics[width=0.31\linewidth]{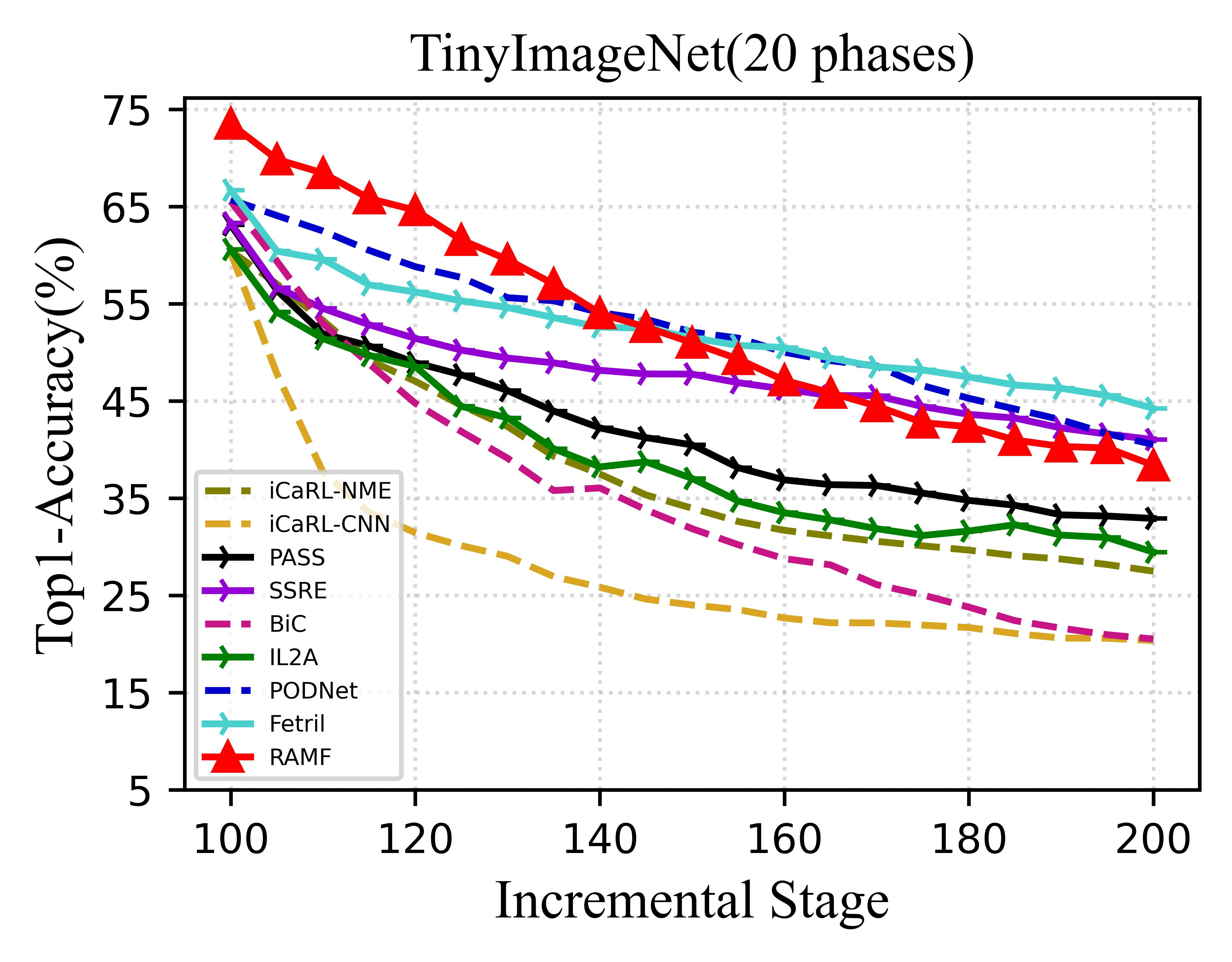}}\\
  \subfloat{	
		\includegraphics[width=0.31\linewidth]{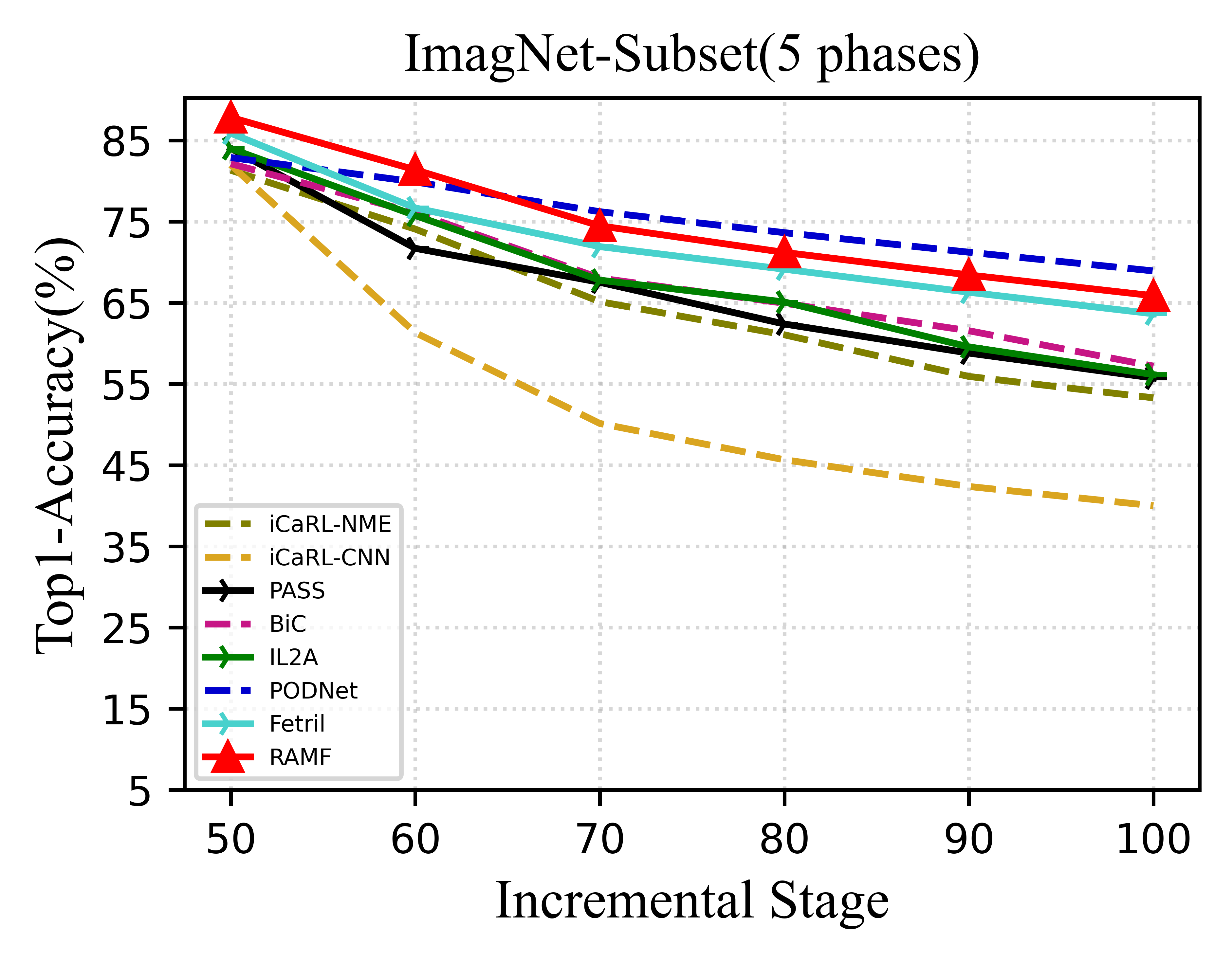}}
	\subfloat{
		\includegraphics[width=0.31\linewidth]{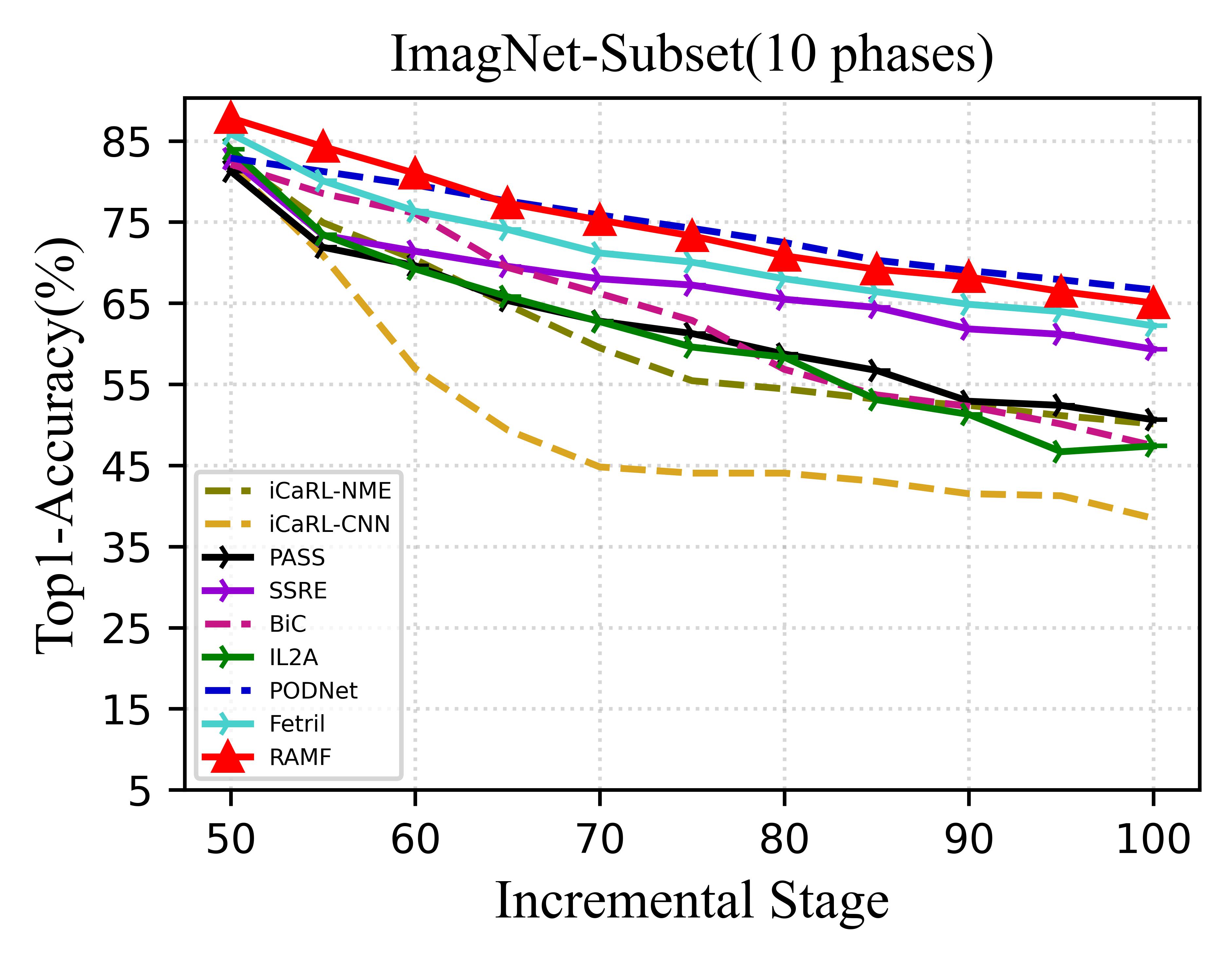}}
	\subfloat{
		\includegraphics[width=0.31\linewidth]{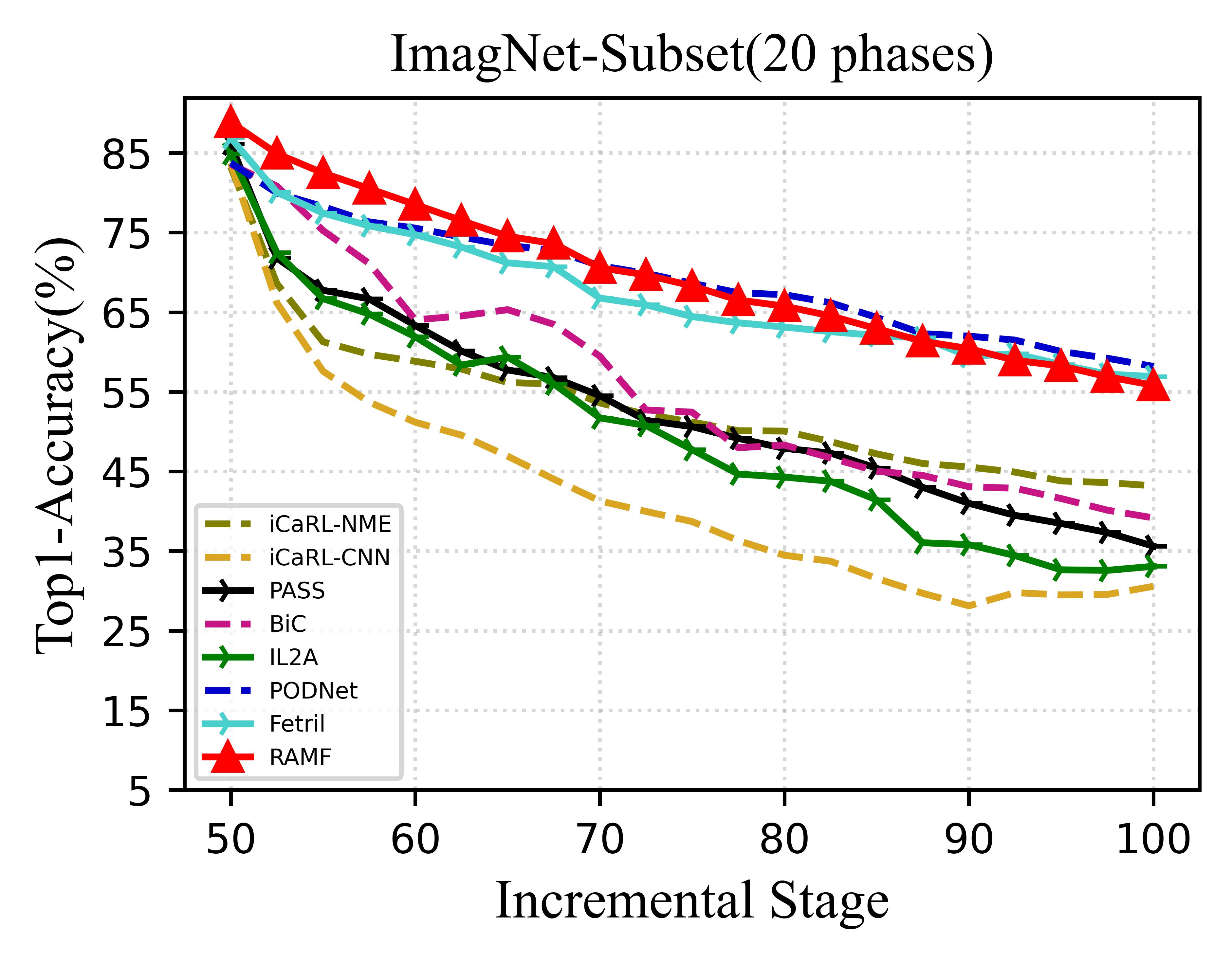}}
	\caption{Accuracy variation over learning stages of different methods on CIFAR-100, TinyImageNet and ImageNet-Subset.}
      \label{accuracy}
\end{figure*}

\subsubsection{Implementation Details} 
Following \cite{il2a,ssre, pass, Fetril}, we use the ResNet-18 initialized randomly as our backbone. In the initial phase, we employ an SGD optimizer with a momentum of 0.9, where the learning rate is 0.1 with cosine annealing scheduling. The weight decay is $5 \times 10^{-4}$. In total, we train the model for 100 epochs for CIFAR-100 and 160 epochs for TinyImageNet and ImageNet-Subset. In each incremental phase, the weight decay is set to $1 \times 10^{-4}$, and the learning rate is adjusted to 0.001 for a total of 50 epochs. We set $\lambda, \mu$ in Eq. \eqref{eq6} as 0.7, 0.3 for CIFAR100 and 1, 1 for both TinyImageNet and ImageNet-Subset. All models are implemented using the Pytorch toolbox and trained on a server with GeForce RTX 3090 GPUs.

\begin{table*}[t]
	\small
	\renewcommand{\arraystretch}{1.4}
	\renewcommand{\tabcolsep}{4pt}
	\centering
	\centering
	\begin{center}
		\caption{Effect of different modules in our method. Accuracy indicates the average accuracy across all classes, and Forgetting for the average forgetting. Same for the remaining Tables.}
		\begin{tabular}{c|ccccccccccc|c|c}
			\hline
			\multirow{1}{*}{Method/Stage}&\textbf{1}  &\textbf{2}  &\textbf{3} &\textbf{4}  &\textbf{5}  &\textbf{6}  &\textbf{7}  &\textbf{8} &\textbf{9} &\textbf{10}  &\textbf{11} &\textbf{Accuracy}&\textbf{Forgetting}\\ \hline
			Baseline     & 81.04 & 75.41 & 70.53  & 66.92  & 63.91 & 59.82  & 57.05  & 53.22  & 53.03   & 50.41  &48.08 &61.76 &32.02\\
			Baseline+Base ClassAug              & 84.14 &  79.00 &  74.45  &  70.50  & 67.18 &  62.78  &  58.65  &  55.58  & 54.20   &  51.48 &48.74 & 64.24 & 26.40\\
			+Mixed Feature(MF)    & 84.14 &  79.00 & 74.15  & 70.73  & 68.17 &  64.78  & 62.03  & 59.28  &  58.12   & 56.01  &52.97 & 66.30 & 22.25\\
			+Auxiliary ClassAug(AC)   & 86.08 & 82.78 & 78.93  & 76.46  & 74.08 & 70.73  & 68.06  & 66.17  & 65.65   & 64.07  &61.43 & 72.22&11.58\\
			\hline
			+MF+AC & \textbf{86.08}  & \textbf{83.09} & \textbf{79.01} &  \textbf{76.58}  & \textbf{73.92}  & \textbf{71.89}  & \textbf{69.83}  & \textbf{67.77}  & \textbf{66.90}   & \textbf{65.75}   & \textbf{63.71}&  \textbf{73.14}&\textbf{10.72}\\ \hline	
		\end{tabular}
		\label{Table}
	\end{center}
\end{table*}
\subsection{Comparison with SOTA}
Since our RAMF is a NECIL method, we mainly compare it to the SOTA NECIL methods. Specifically, we select IL2A \cite{il2a}, PASS \cite{pass}, SSRE \cite{ssre}, and Fetril \cite{Fetril}. Besides, some classical and latest exemplar-based approaches are also compared, including iCaRL \cite{icarl}, BiC \cite{Bic}, PODNet \cite{PODNET} and FOSTER \cite{foster}. For a fair comparison, the backbone of all models is ResNet-18. 

\subsubsection{Average Accuracy} 
As illustrated in Table \ref{avgacc}, our proposed method RAMF outperforms all the compared NECIL methods by a large margin, which validates its effectiveness. Specifically, compared to IL2A, which only employs Base classAug, our method obtains an average improvement of 11.59\%, 12.93\% and 13.28\% on CIFAR100, TinyImageNet and ImageNet-Subset respectively. This validates that the auxiliary class augmentation can improve the learning ability of a model. Our RAMF surpasses the best NECIL method Fetril \cite{Fetril} on all settings largely. For instance, the improvement on CIFAR-100 is over 7\%. Furthermore, we achieve comparable performance to the latest exemplar-based method FOSTER \cite{foster} in most cases. And the margin is large on CIFAR-100.

\begin{table*}[t]
	\setlength{\tabcolsep}{5pt}
	\small
	\renewcommand{\arraystretch}{1.4}
	\centering
	\caption{Performance of different random auxiliary classes augmentation on CIFAR-100 (11 Phases). Base: Base ClassAug, R: Rotation, P: Color Permutation, O:CutOut.}
	\begin{tabular}{c|ccc|ccccccccccc|c|c}
		\hline		
	    \textbf{Base} & \textbf{R} &{\textbf{P}} &{\textbf{O}} &\textbf{1} &\textbf{2}  &\textbf{3}  &\textbf{4}  &\textbf{5}  &\textbf{6} &\textbf{7}  &\textbf{8} &\textbf{9}  &\textbf{10}  &\textbf{11}  &\textbf{Accuracy} &\textbf{Forgetting}  \\ 
		
		\hline
		$\surd$                   & & &    					& 83.98  & 79.49&76.70 &73.36&69.78&66.25&63.87&60.88&59.42&57.48&54.14&67.76&23.18  \\ 
		$\surd$                   & $\surd$ &         &  		& 86.02  & 82.29 &78.48 &76.21&73.81&70.32&67.81&65.27&63.98&62.72   &60.11   &71.55   &13.41 \\  
		$\surd$					  & $\surd$  &$\surd$ &   		& 86.02    & 82.54  &78.75 &76.55&73.88&70.20&68.51&66.34&65.02&63.85&61.92&72.13&13.03 \\ 
		$\surd$                   & $\surd$  &$\surd$ &$\surd$ &  \textbf{86.08}    & \textbf{83.09}   &\textbf{79.01}&\textbf{76.58}&\textbf{73.92}&\textbf{71.89}&\textbf{69.83}&\textbf{67.77}&\textbf{66.90}&\textbf{65.75}&\textbf{63.71}&\textbf{73.14}&\textbf{10.72}  \\ 
		\hline
	\end{tabular}
	\label{ablation}
\end{table*}

\begin{table*}[t]
	\small
	\renewcommand{\arraystretch}{1.4}
	\renewcommand{\tabcolsep}{5.8pt}
	\centering
	\caption{Influence of different weights for random auxiliary classes augmentation on CIFAR-100 (11 Phases).}
	\begin{tabular}{c|ccccccccccc|c|c}
		\hline
		\multirow{1}{*}{\textbf{Weight/Stage}}  &\textbf{1}  &\textbf{2}  &\textbf{3} &\textbf{4}  &\textbf{5}  &\textbf{6}  &\textbf{7}  &\textbf{8} &\textbf{9} &\textbf{10}  &\textbf{11} &\textbf{Accuracy} &\textbf{Forgetting} \\ 
		\hline
		\textbf{1:1:1}  &  \multicolumn{1}{c}{84.80 }  & \multicolumn{1}{c}{80.92}  & \multicolumn{1}{c}{77.56}    &74.92&72.54&68.62&67.33&65.18&63.92&62.70&60.21 &70.79&11.74         \\ 
		\cline{1-1}
		\textbf{4:1:1}  & \multicolumn{1}{c}{85.62}   & \multicolumn{1}{c}{81.54}   & \multicolumn{1}{c}{78.15}      &75.80&73.15&69.64&67.38&65.57&64.71&63.16&60.64 &71.39&12.67               \\
		\cline{1-1}
		\textbf{8:1:1}  & \multicolumn{1}{c}{\textbf{86.08}}   & \multicolumn{1}{c}{\textbf{83.09}}   & \multicolumn{1}{c}{79.01}     &\textbf{76.58}&73.92&\textbf{71.89}&\textbf{69.83}&\textbf{67.77}&\textbf{66.90}&\textbf{65.75}&\textbf{63.71} &\textbf{73.14}& \textbf{10.72}        \\ 
		\cline{1-1}
		\textbf{1:8:1}  & \multicolumn{1}{c}{85.70}   & \multicolumn{1}{c}{81.60}   & \multicolumn{1}{c}{77.61}     &75.63&72.88&69.45&67.17&64.96&63.85&62.36&59.96 &71.01& 13.45        \\
		\cline{1-1}
		\textbf{1:1:8}  & \multicolumn{1}{c}{84.40}   & \multicolumn{1}{c}{79.94}   & \multicolumn{1}{c}{76.01}     &74.03&71.34&67.32&64.68&61.81&61.27&59.37&56.70 &68.79& 14.17        \\
		\cline{1-1}
		\textbf{12:1:1}  & \multicolumn{1}{c}{86.00}  & \multicolumn{1}{c}{82.35}   & \multicolumn{1}{c}{\textbf{79.08}}      &76.41&\textbf{74.12}&70.92&68.75&66.20	&65.44&63.53&61.22 &72.18&13.16        \\ 
		\cline{1-1}
		\hline
	\end{tabular}
	\label{weight}
\end{table*}

\begin{table*}[t]
	\small
	\renewcommand{\arraystretch}{1.4}
	\renewcommand{\tabcolsep}{5.pt}
	\centering
	\caption{Performance of randomly select one auxiliary classes augmented method (Random Learning) and joint training of all auxiliary classes augmentations (Joint Learning) on CIFAR-100 (11 Phases).}
	\begin{tabular}{c|ccccccccccc|c|c}
		\hline                                  
		\multirow{1}{*}{\textbf{Method}}    &\textbf{1} &\textbf{2}&\textbf{3}&\textbf{4}&\textbf{5}  &\textbf{6} &\textbf{7}  &\textbf{8}  &\textbf{9}&\textbf{10}  &\textbf{11} &\textbf{Accuracy} &\textbf{Forgetting} \\ 
		\hline
		\textbf{Joint Learning} &  84.50  & 78.05  & 75.01  &72.12  &70.24  &67.97  &64.28  &62.80 
        &60.38  &59.41  &56.57 &68.30&12.62                      \\ \cline{1-1}
		\textbf{Random Learning} & \textbf{86.14}  & 82.52  & 78.93 &76.32&73.20&69.05&66.55&63.38&62.54&60.46&58.49 &70.69&13.51             \\ \cline{1-1}
        \textbf{Ours}  & 86.08   & \textbf{83.09}   & \textbf{79.01} &\textbf{76.58}&\textbf{73.92}&\textbf{71.89}&\textbf{69.83}&\textbf{67.77}&\textbf{66.90}&\textbf{65.75}&\textbf{63.71} &\textbf{73.14}& \textbf{10.72}  \\ \cline{1-1}
        \hline
	\end{tabular}
	\label{all_phase}
\end{table*}

\subsubsection{Interplay of Forgetting and Intransigence} 
Since stability and plasticity influence each other, we further investigate the balance between plasticity and stability for different methods from the interaction of forgetting and intransigence. The results of different methods are shown in Figure \ref{for-intra}, where the model points closer to the bottom of the image indicate less forgetting and higher stability, and those closer to the left indicate less intransigence and higher plasticity. In this figure, most methods are easily biased towards new tasks leading to catastrophic forgetting. To mitigate catastrophic forgetting, we proposed the class augmentation and mixed features to improve the generalization. Although it may limit the performance on new tasks, we strike a better balance between stability and plasticity. 

\subsubsection{Accuracy Curve} 
To compare the accuracy variation at each stage, we show the detailed accuracy curves for different partitions of three datasets in Figure \ref{accuracy}. In this figure, we obtain the best initial accuracy, which shows the random auxiliary classes augmentation in the initial stage can significantly improve the model's generalization. At almost all stages, we achieve superior accuracy on different datasets, showing that mixed features and auxiliary classes are applicable in different incremental scenarios.

\subsection{Ablation Study}
\subsubsection{Influence of Different Modules} 
To investigate the effectiveness of each module, we have conducted ablation studies on CIFAR-100, whose results are in Table \ref{Table}. The baseline consists of noisy prototype, cosine normalization and knowledge distillation. From this Table, we can get three points. 1) Adding the Base ClassAug reduces forgetting by 4.39\%, demonstrating that extra classes can improve model generalization. 2) Replacing new features with mixed features further reduces forgetting by 4.15\%. Moreover, when the incremental stage increases, the accuracy improvement by using mixed features also becomes bigger, which validates the importance of limiting new features learning in long-term incremental learning. 3) Adding auxiliary class augmentation significantly improves the average accuracy by 7.45\% and reduces forgetting by 16.05\%, which verifies the advantage of Auxiliary ClassAug over Base ClassAug. More augmented images and extra classes in the first stage help the model to resist true forgetting in later incremental stages. Finally, combining all modules leads to the best results.

\begin{figure*}[t] 
	\centering 
	\subfloat[PASS]{	
		\includegraphics[width=0.23\linewidth]{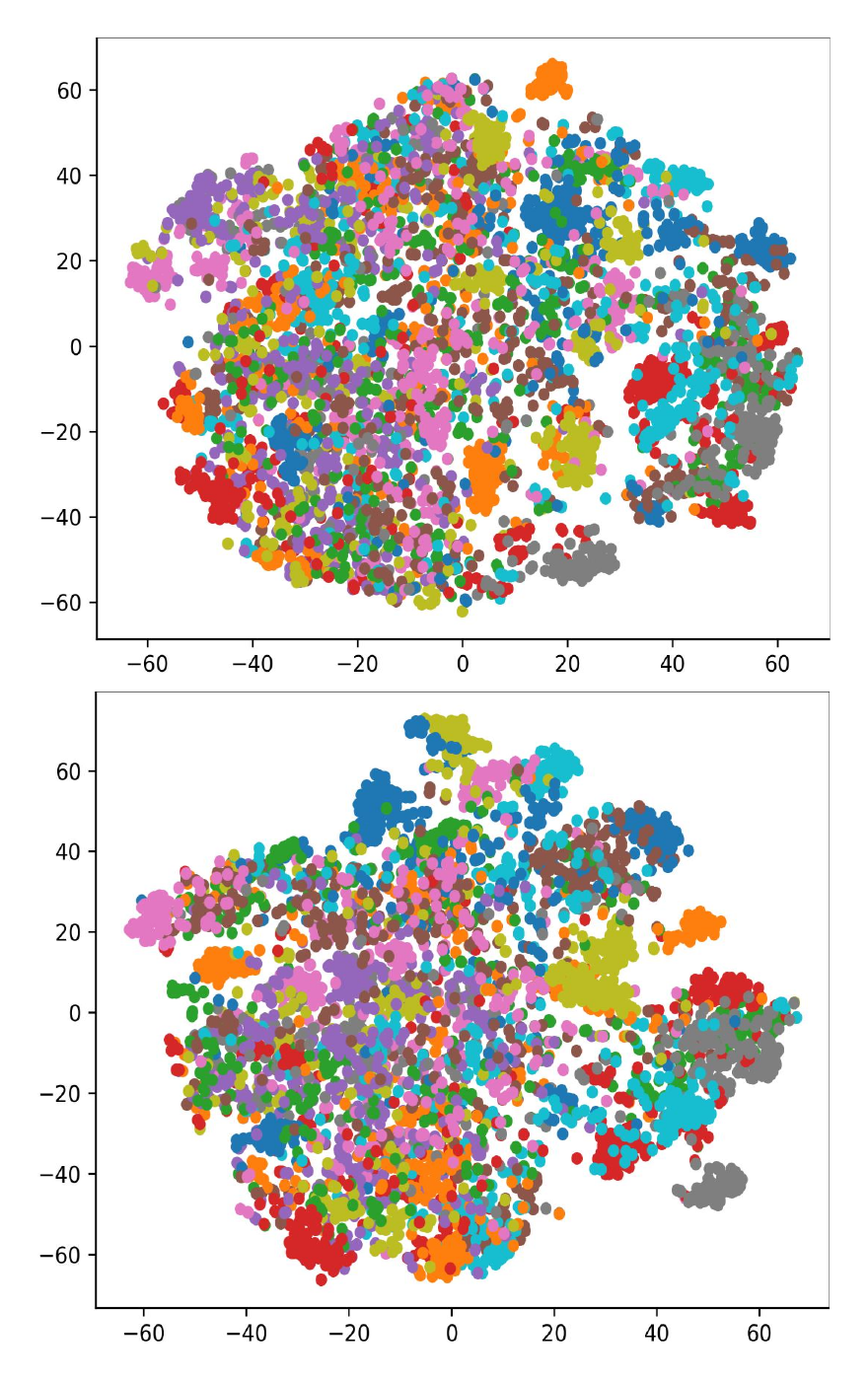}
		\label{PASSTsne}
	     }
	\subfloat[iCaRL]{
		\includegraphics[width=0.23\linewidth]{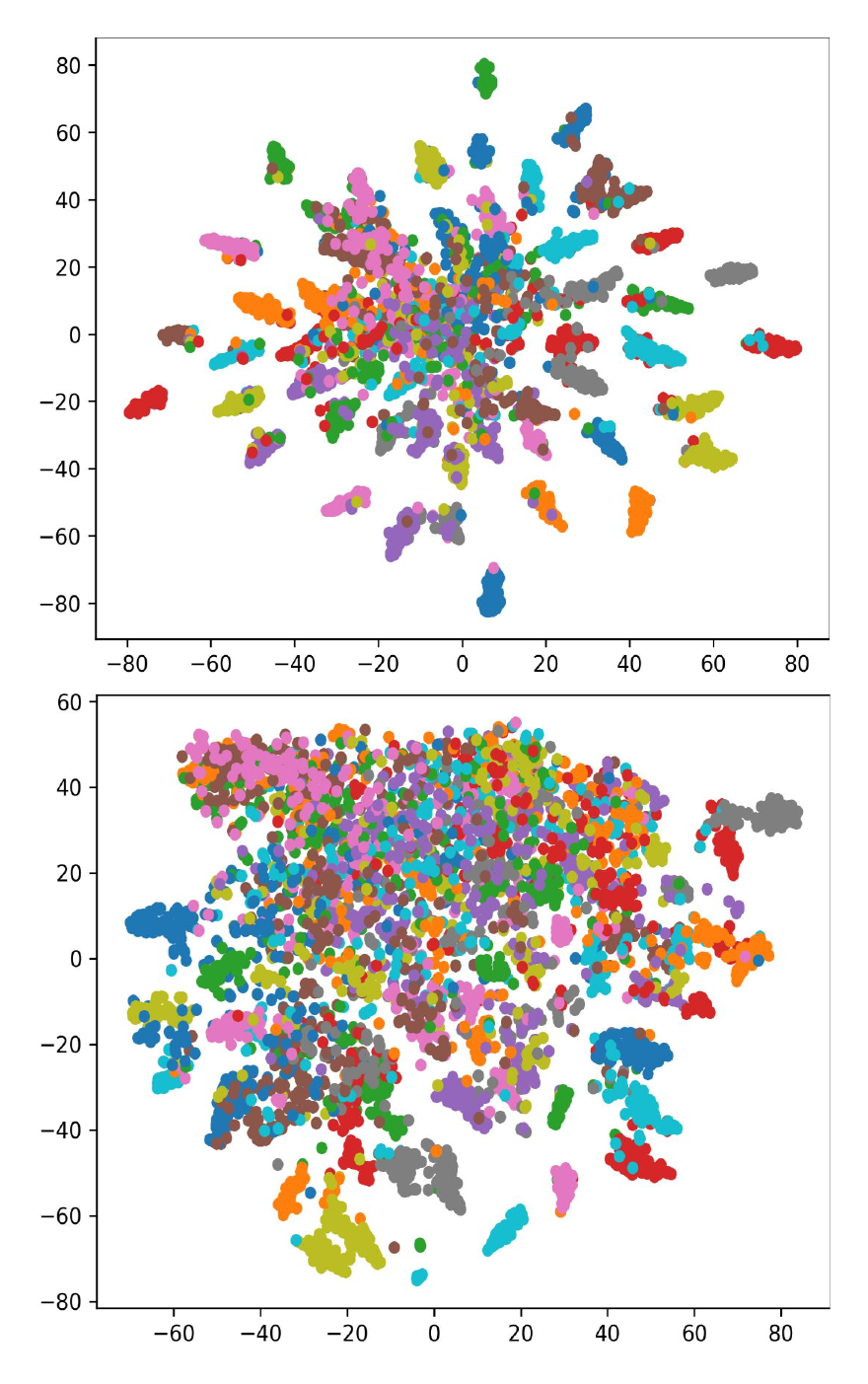}
		\label{iCaRLTsne}
	}
	\subfloat[IL2A]{
		\includegraphics[width=0.23\linewidth]{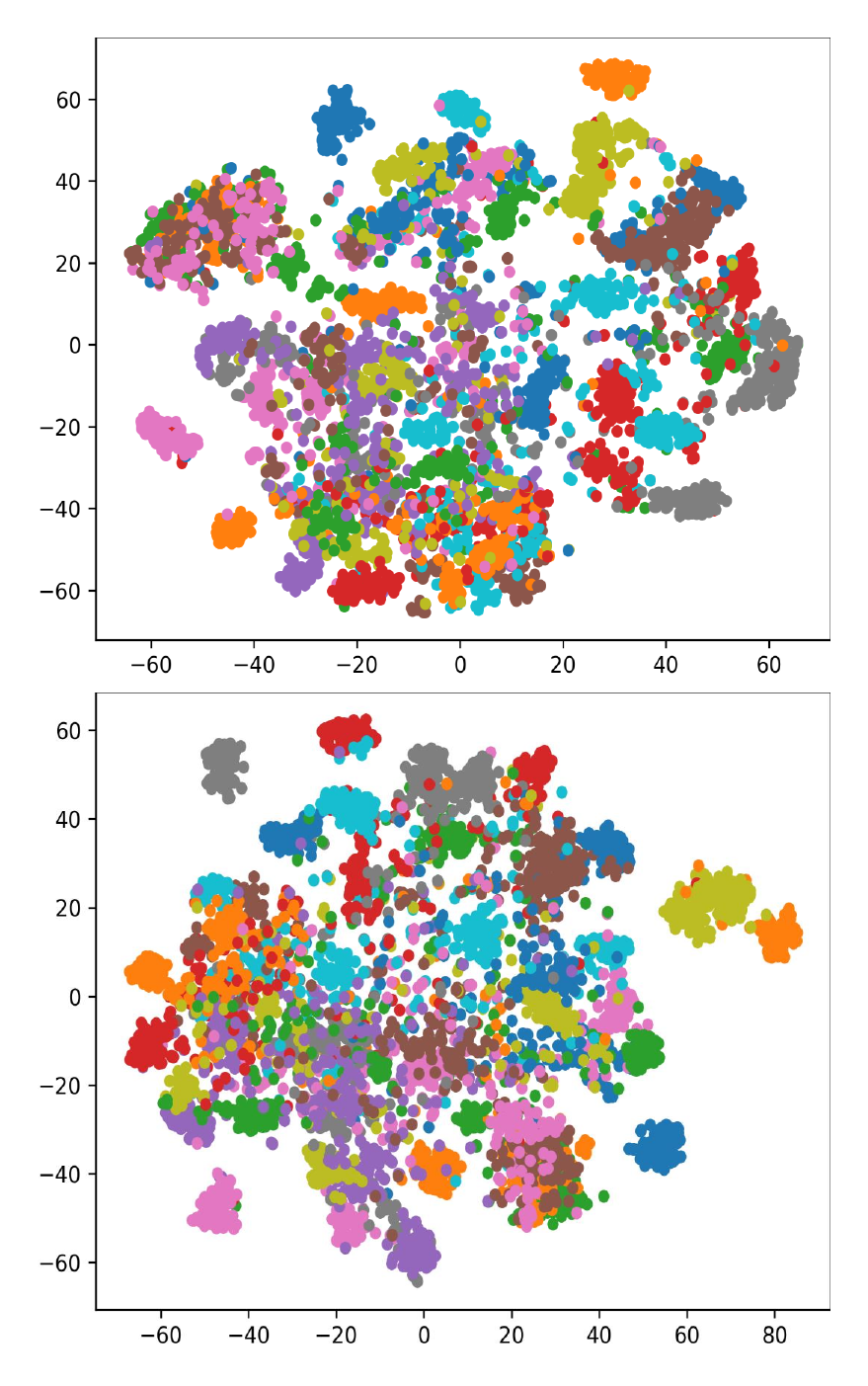}
		\label{IL2ATsne}
	}
	\subfloat[RAMF]{
		\includegraphics[width=0.23\linewidth]{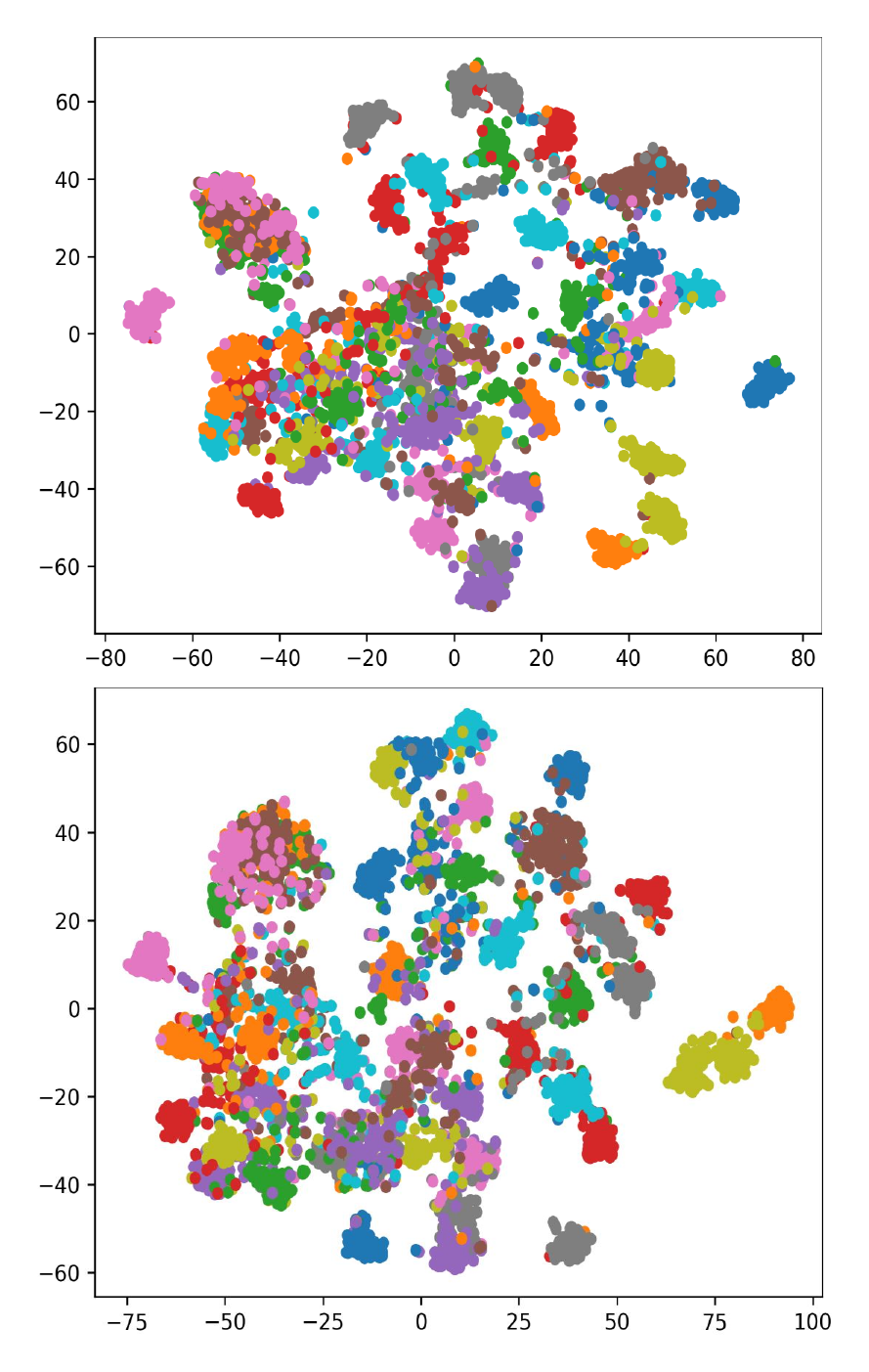}
		\label{RAMFTsne}
	}
	\caption{t-SNE visualization of features space for the 50 base classes of different methods on CIFAR-100. The upper figures represent the feature space after learning 50 base classes. The lower figures represent the feature space for the base 50 classes after learning all classes.}
	\label{tsne}
\end{figure*}

\begin{figure}[t]
        \vspace{-0.1cm}
	\centering  
	\subfloat[iCaRL]{
		\includegraphics[scale=0.3]{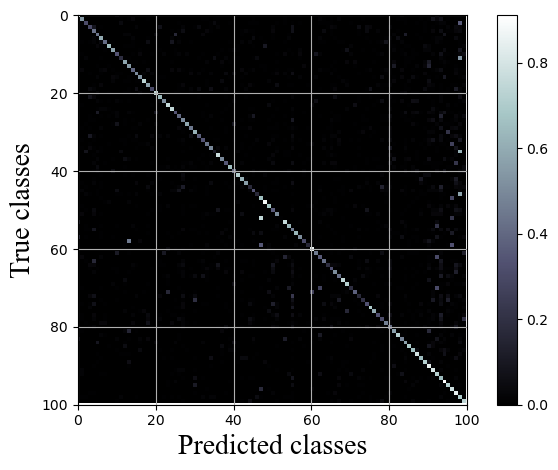}}\subfloat[IL2A]{
		\includegraphics[scale=0.3]{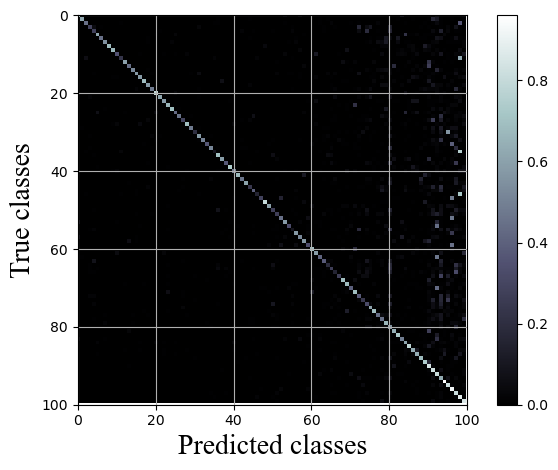}}\\
	\subfloat[PASS]{
		\includegraphics[scale=0.3]{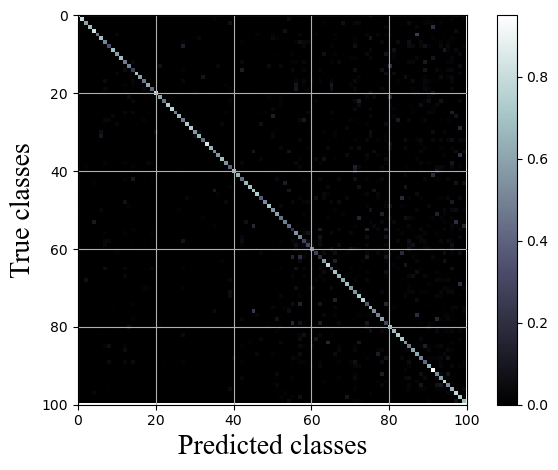}}\subfloat[RAMF]{
		\includegraphics[scale=0.3]{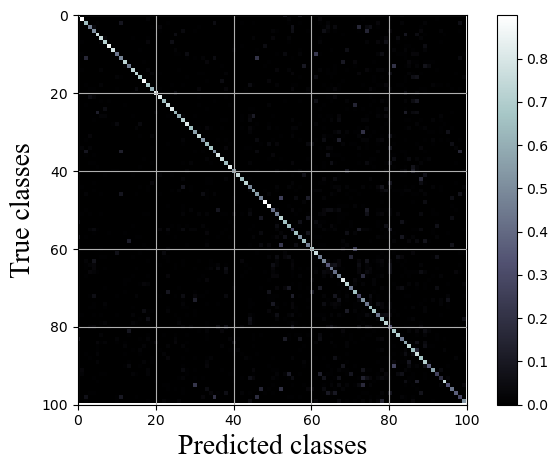}}
	\caption{Comparison of confusion matrix of iCaRL, IL2A, PASS and RAMF.}
	\label{matrix}
\end{figure}
\subsubsection{Influence of Auxiliary Classes Augmentation} 
To explore the influence of configuration for random auxiliary class augmentation, we performed experiments on CIFAR-100. As illustrated in the first two rows of Table \ref{ablation}, the introduction of rotation increases the accuracy by 3.79\% and reduces forgetting by 9.77\%. Combining the rotation, color permutation and cutout obtain the best performance. All these scores demonstrate the effectiveness of auxiliary classes in improving the model's generalization as well as resisting forgetting.

Table \ref{weight} gives the influence of weights for rotation, cutout and color permutation. We can see the best performance is obtained when the weights are 8:1:1. This declares the importance of rotation, which can make the model grasp more fined features to discriminate these similar classes. Besides, optimizing the model towards one auxiliary class is better than optimizing it equally among several auxiliary classes. 

As stated before, we use different auxiliary class augmentation methods for initial and incremental stages. To validate its influence, we have conducted some experiments, whose results are shown in Table \ref{all_phase}. In random learning, only one auxiliary augmented method is selected randomly in both two stages. In joint learning, all three components are used in both two stages. In this Table, random learning obtains large accuracy than joint learning, which indicates that learning too many additional classes simultaneously will degrade performance. Moreover, our method achieves the best performance. In our opinion, we can train a robust feature space by introducing randomness in the initial phase, while the use of fixed auxiliary class augmentation in the incremental phase minimizes the perturbation to the feature space generated by the new task.

\subsubsection{Mixed Features Under Different Class Augmentation}
To investigate the effect of the proportion of new and old features in mixed features in various class augmentation scenarios, we conducted experiments with different weights on CIFAR-100. From the table \ref{mix_feature}, we can see that Base classAug improves the overall performance but has limited improvement in the last stage accuracy. In contrast, the random auxiliary class augmentation drastically improves all the scores, which illustrates the effectiveness of our approach in improving the model's transferability. Without class augmentation, there are few transferable features in the original classes. Learning the current stage of transferable features with new features will lead to more negative effects due to the disturbance of feature space than the transferability brought by these transferable features, so only using the features extracted by the old model can best avoid bias. With the addition of Base ClassAug, the transferability of extracted features increases compared to the original model. However, the increase is small, so using a smaller proportion of new features to fit the additional transferable features can better maintain the stability of the feature space. Compared with this, the transferability of the model increases significantly when random auxiliary class augmentation is added. Therefore, when occupying a higher proportion, the new features facilitate the model to learn the transferable features at the current stage and improve the model's generalization across tasks.
\begin{table*}[t]
	\small
	\renewcommand{\arraystretch}{1.4}
	\renewcommand{\tabcolsep}{6pt}
	\centering
	\label{Mixed Feature}
	\caption{Influence of the proportion of new and old features in Mixed Feature on CIFAR-100 (11 Phases). Init and Last represent the initial and last stage accuracy, respectively, and Avg and Forg represent the average accuracy and the average forgetting, respectively. $\alpha$ is $\sqrt{\frac{ N_{new}}{N_{old}}}$, $N_{new}$ and $N_{old}$ represent the number of novel classes and learned classes.}
	\begin{tabular}{ccc|cccc|cccc|ccccc}
		\hline
  &\multirow{2}{*}{\textbf{$f_{t-1}$}} &\multirow{2}{*}{{$f_{t}$}} 
  &\multicolumn{4}{c|}{Without ClassAug}&\multicolumn{4}{c|}{+Base ClassAug} &\multicolumn{4}{c}{+Random Auxiliary ClassAug}\\
  &&&\textbf{Init}&\textbf{Last}&\textbf{Avg}&\textbf{Forg}
  &\textbf{Init}&\textbf{Last}&\textbf{Avg}&\textbf{Forg} &\textbf{Init}&\textbf{Last}&\textbf{Avg}&\textbf{Forg}
  \\ 
		
		\hline
		&\textbf{0}   &\textbf{1}   
            &81.42  &47.85 &61.93 &33.51       
            &84.14  &48.74  &64.24   &26.40
            &86.08  &61.42  & 72.22   &11.58\\ 
		
		&\textbf{0.3}   &\textbf{0.7}  
            &81.42 &48.48 &62.09 &33.10 
            &84.14 &50.26 &65.43 &26.09
            & 86.08 &\textbf{63.71}  &\textbf{73.14}&10.72\\
		
		&\textbf{0.5}   &\textbf{0.5}   
            &81.42&49.89 &62.67 &32.68    
            &84.14 &52.27 &65.70 &24.08
            & 86.08 &62.03  &72.46&11.29\\ 
		
		&\textbf{0.7}   &\textbf{0.3} 
            &81.42&51.01&62.92 &31.15 
            &84.14 &52.97 &66.30 &\textbf{22.25}
            & 86.08 &62.23  &72.56&\textbf{10.44}\\ 
		
	    &\textbf{1}   &\textbf{0}  
            &81.42 &\textbf{51.76} &\textbf{63.62} &\textbf{29.90} 
            &84.14 &52.48 &62.22 &23.80
             & 86.08 &62.22  &72.57&11.79\\ 
	    
	    &\textbf{$\bm{1-\alpha}$} &\textbf{$\bm{\alpha}$}   
            &81.42&50.02 &62.82 &31.89 
            &84.14&\textbf{52.98} &\textbf{66.33} &23.03
            &86.08 &61.98  &72.38&11.85\\ 
		\hline
	\end{tabular}
	\label{mix_feature}
\end{table*}
\subsection{t-SNE of Features}
To show the discrepancy of decision boundaries before and after incremental learning, we visualize the feature distribution of 50 base classes for different methods using t-SNE in Figure \ref{tsne}. All methods adopt the same class division, which first learn 50 base classes and then learn the remaining 50 classes in 5 incremental stages. As seen from Figure \ref{PASSTsne}, the decision boundaries of PASS mixed as a mass despite the self-supervised learning involved. For iCaRl, a severe drift occurs in the feature space after incremental learning in Figure \ref{iCaRLTsne}. We argue this is because it occurs over-fitting when learning the first 50 classes. Due to the storage of old data, part of the decision boundary remains stable. However, most areas are mixed up, which illustrates how over-fitting can cause the forgetting of old knowledge after learning new tasks. Since IL2A introduces a lot of additional classes, it has more stable decision boundaries. However, the large number of additional classes makes it difficult for the model to learn the original features. Meanwhile, it does not solve the problem of decision boundary drift caused by new classes. Therefore, the feature space of IL2A is not sufficiently separated as shown in Figure \ref{IL2ATsne}. In contrast, our proposed RAMF produces better boundaries in both the initial and final stages in \ref{RAMFTsne}, which also show the effectiveness of our proposed modules.

\subsection{Comparison of Confusion Matrix} 
To evaluate the performance of the model for the old and new classes, we compared the confusion matrix of iCaRL, IL2A, PASS, and RAMF in Figure \ref{matrix}. The diagonal represents correct classification while the off-diagonal represents incorrect classification. As shown in this Figure, both iCaRL and IL2A tend to predict old classes as new classes in the later stages. They obtain much higher accuracy on the last task than that on the old tasks, which means the features of new class dominate the feature space. In contrast, PASS and our RAMF show similar accuracy on the new and old classes. Furthermore, RAMF obtains higher accuracy rate than PASS, which results in better overall performance.

\subsection{Transferability and Forgetting}
\begin{figure}[t] 
	\centering 
	\subfloat[New task accuracy]{	
		\includegraphics[width=0.45\linewidth]{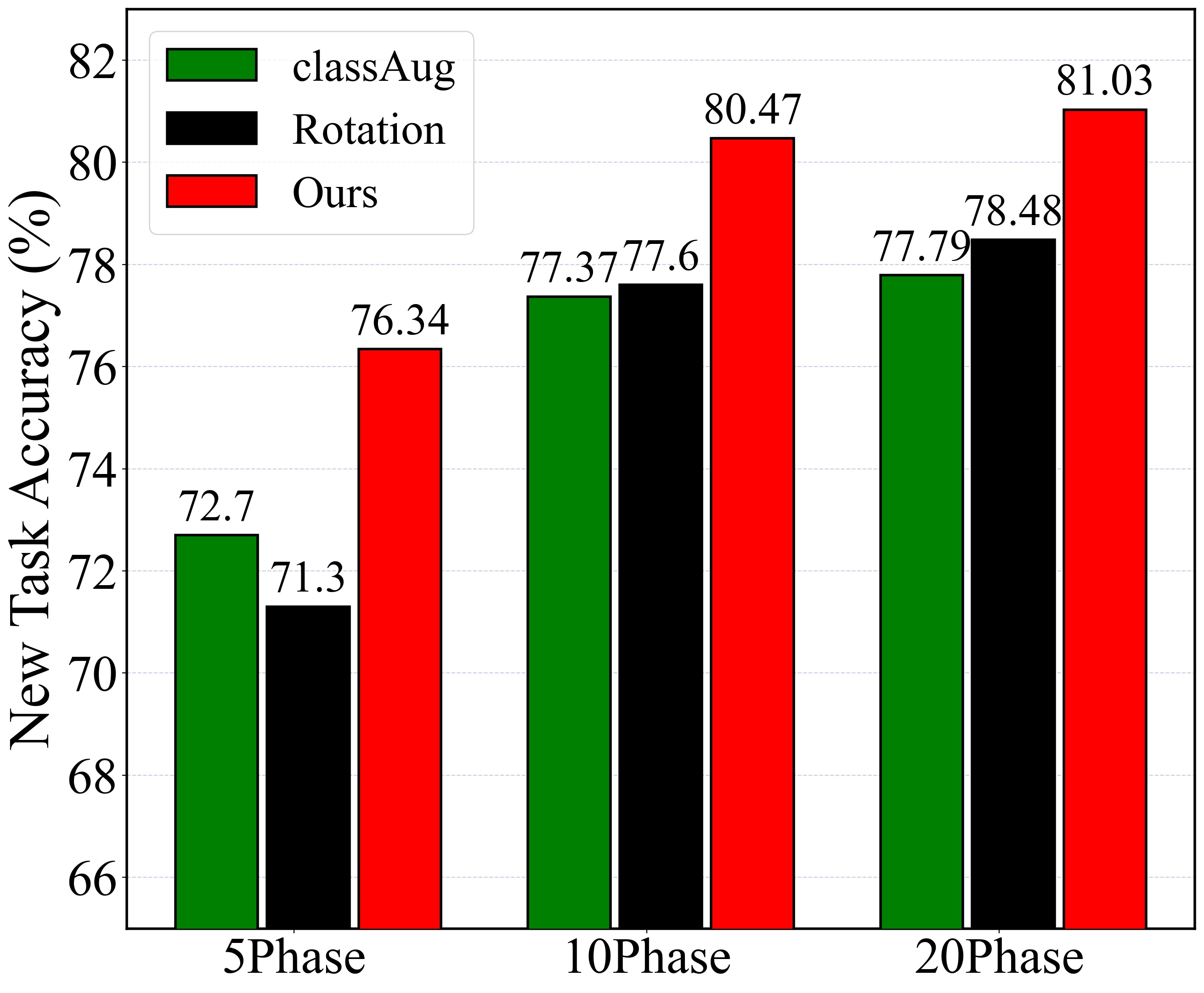}
		\label{bar_accuracy}
	     }
	\subfloat[Forgetting]{
		\includegraphics[width=0.45\linewidth]{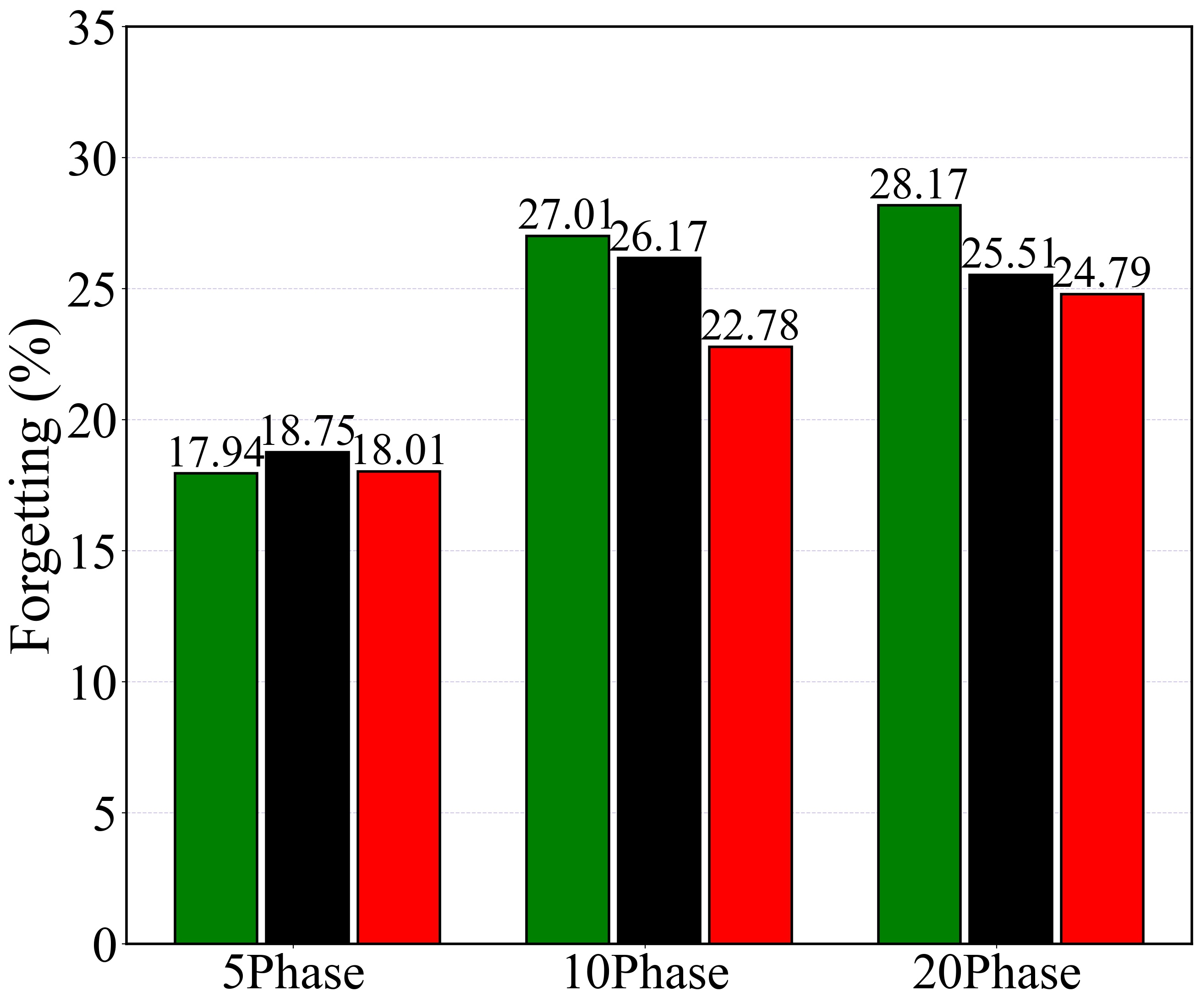}
		\label{bar_forgetting}
	}
 \caption{Performance of different augmentation methods on new task accuracy and forgetting in CIFAR-100.}
	\label{bar}
\end{figure}

To explore the impact of different class augmentation methods on transferability to new classes and forgetting of old knowledge, we conducted various experiments on the CIFAR-100 dataset with different numbers of new classes. When the incremental phase is small, classAug has far more transferable features to learn than rotation since there are more augmented classes. Even in this case, as shown in Figure \ref{bar_accuracy}, its accuracy is only 1.4 points higher than rotation, which suggests that the mixed features affect the recognition of the original classes. When there are few new classes, the transferability of classAug even become worse than rotation. In contrast, our method obtains the best transferability regardless of the number of new classes. From Figure \ref{bar_forgetting}, we can observe that our method has less forgetting than other methods, given a better transferability. This suggests that the transferable features are beneficial in enabling the model to transfer between tasks.

\subsection{Memory Analysis} 
Low memory requirement is desirable for incremental learners due to the huge number of classes in real world. Since nearly all methods employ the ResNet-18 as backbone, we will not consider it in the following analysis. Most of the current NECIL methods (IL2A, PASS, SSRE and Fetril) use prototypes to represent old classes, each of which contains 512 parameters. Therefore, these methods including ours just need 51.2K memory space for prototypes for a dataset of 100 classes. Note that IL2A requires additional storage to store the class covariance matrix, i.e. 512x512 parameters per class. In contrast, exemplar-based methods need additional memory for exemplars, whose size depends on the dataset. Since most replay-based methods store 20 samples per class, they require $100\times20\times32\times32\times3$ parameters, or about 9.2M of memory for CIFAR-100. For ImageNet-SubNet, the number of parameters will be $100\times20\times224\times224\times3$, which is about 301M. From the above analysis, we can see the memory advantage of our NECIL method.

\section{Conclusion}
In this paper, we proposed the RAMF, a new approach for non-exemplar class-incremental learning. For each input batch, we perform Base ClassAug and Random Auxiliary ClassAug to generate extra samples and labels. Combining these two augmentation methods allows the model to obtain a robust feature space at the initial stage, which will relieve forgetting during real incremental learning. Furthermore, we provide new insights for resisting catastrophic forgetting. We use mixed features to replace new class features to prevent the model from biasing towards new classes. The mixed features can avoid learning too many new features simultaneously, which may cause a shift of old features. In other words, the mixed feature is beneficial to keep the stability of the feature space. Extensive experiments on popular datasets confirm the effectiveness of our approach. Especially, the performance of this new method can be compared to exemplar-based methods, which can help to avoid the privacy concern caused by storing samples.

Despite the superior performance of the combination of random auxiliary class augmentation and mixed features, they limit the model's ability to learn new tasks to a certain degree. This is the well-known plasticity-stability dilemma in continual learning scenarios. How to improve the model's plasticity will be considered in the future.

{\small
\bibliographystyle{IEEEtran}
\bibliography{egbib}
}
\end{document}